\documentclass[]{dataflow}

\usepackage[toc,page,header]{appendix}

\usepackage{minitoc}
\usepackage{multirow}     
\usepackage{multicol}     
\usepackage{booktabs}     
\usepackage{colortbl}     
\usepackage[table]{xcolor} 
\usepackage{makecell}     
\usepackage{graphicx}     
\usepackage{subcaption}
\usepackage{amssymb}
\usepackage{makecell}
\usepackage{minted}
\usepackage{fontawesome5}
\usepackage{amsfonts}
\title{DataPrep-Bench: Benchmarking LLMs as Training Data Preparators}

\definecolor{lightyellow}{RGB}{255, 255, 204}  
\definecolor{lightgreen}{RGB}{204, 255, 204}   

\newcommand{\pnum}[1]{\cellcolor{lightyellow}#1}  
\newcommand{\nnum}[1]{\cellcolor{lightgreen}#1}   

\author[*, \dagger]{Hao Liang}
\author[*]{Qifeng Cai}
\author[*]{Yibo Lin}
\author[*]{Jianzhuo Du}
\author[]{Qifeng Xia} 
\author[]{Sizhe Qiu} 
\author[]{Linzhuang Sun} 
\author[]{Meiyi Qiang} 
\author[]{Zhaoyang Han} 
\author[]{Xiaochen Ma} 
\author[]{Bohan Zeng} 
\author[]{Ruichuan An} 
\author[]{Conghui He}
\author[]{Wentao Zhang}

\affiliation[]{$^{1}$Peking University, $^{2}$Institute for Advanced Algorithms Research, Shanghai, $^{3}$OriginHub Technology, $^{4}$Zhongguancun Academy}

\abstract{
The quality of training data fundamentally determines the capabilities of large language models (LLMs), yet no unified benchmark exists to measure how well LLMs, agents, and data-centric workflows actually prepare training data end to end. We view LLM-driven data preparation as comprising two complementary capabilities: \textbf{data construction}, which transforms raw sources into supervised training data, and \textbf{data quality evaluation}, which predicts the training value of candidate datasets before downstream training; throughout, ``quality'' refers to downstream training utility rather than surface-level textual properties. We introduce \textbf{DataPrep-Bench}, the first unified benchmark that jointly evaluates both capabilities under a shared downstream-grounded protocol over six domains and multiple base models. For \textbf{data construction}, methods consume identical raw sources and are scored by fine-tuning a base model on their outputs jointly with Dolly-15k; alongside this track we release \textbf{Data-Construction-Skill}, a skill-guided agent that lifts the Dolly-only baseline by nearly 20 points absolute on Llama-3.1-8B Finance and is competitive with the strongest agent- and DataFlow-based methods in knowledge-extraction-dense domains. For \textbf{data quality evaluation}, scoring functions are scored by Pearson correlation with downstream performance on a shared candidate pool; we release the \textbf{Distributional Alignment Score (DAS)}, a distribution-based evaluator that uses MMD between a candidate dataset and a domain proxy. DAS attains the strongest cross-model correlation in four of six domains and is the only metric clearing $r>0.70$ simultaneously in Math, Science, and Medical, outperforming existing quality-, diversity-, and heuristic-based evaluators. DataPrep-Bench provides a unified, downstream-grounded framework for measuring progress on both capabilities as co-equal targets of LLM-driven data preparation.

}

\date{\today}

\def\emailicon{\raisebox{-1.5pt}{\includegraphics[height=1.05em]{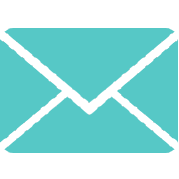}}}
\def\githubicon{\raisebox{-1.5pt}{\includegraphics[height=1.05em]{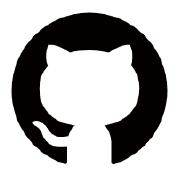}}}
\def\huggingfaceicon{\raisebox{-1.5pt}{\includegraphics[height=1.05em]{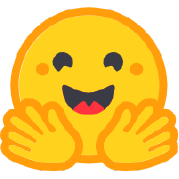}}}

\contribution[*]{Equal Contribution}
\contribution[\dagger]{Project Leader}

\newcommand{\sourcelink}{https://github.com/haolpku/Data-Preparation-Bench}
\newcommand{\datalink}{https://huggingface.co/datasets/lhpku20010120/Data-Prep-Bench}

\checkdata[ \githubicon \hspace{0.3em} Source Code ]{ \url{\sourcelink} }
\checkdata[ \huggingfaceicon \hspace{0.3em} Datasets \& Demo ]{ \url{\datalink} }
\checkdata[ \faFile \hspace{0.57em} Benchmark Website ]{ \url{https://datapreparationbench.github.io/} }
\checkdata[ \emailicon \hspace{0.3em} Correspondence ]{\email{wentao.zhang@pku.edu.cn}}

\begin{document}
\maketitle

\renewcommand{\thefootnote}{\fnsymbol{footnote}} 
\setcounter{footnote}{0}


\renewcommand{\thefootnote}{\arabic{footnote}}
\pagestyle{fancy}
\fancyhf{}
\fancyhead[L]{DataPrep-Bench}
\fancyhead[R]{\thepage}

\newpage
\tableofcontents
\newpage

\section{Introduction}

The performance of large language models (LLMs) is fundamentally governed by the quality, diversity, and scale of their training data~\cite{kaplan2020scaling}, and carefully curated data can match or beat much larger noisy corpora~\cite{zhou2023lima, gunasekar2023textbooks}. As naturally occurring corpora fail to keep up with demand, the community increasingly uses LLMs, agents, and data-centric workflows to produce training data themselves.

We refer to this paradigm as \emph{LLM-driven data preparation} and decompose it into two capabilities. \textbf{Data construction} transforms raw, non-trainable sources (domain books, technical manuals, web dumps) into supervised training data, as in Self-Instruct~\cite{wang2023self}, Evol-Instruct~\cite{xu2304wizardlm}, and UltraChat~\cite{UltraChat}. \textbf{Data quality evaluation} predicts, before training, which candidate datasets will improve a downstream model, through signals such as influence estimation, distributional analysis, or LLM-as-a-judge scoring~\cite{xia2024less}. Although both capabilities have a fast-growing literature, cross-method comparison is largely anecdotal: existing works use disparate sources, base models, and downstream benchmarks, so it is unclear which construction strategies actually yield useful training data and whether any given quality score is genuinely predictive of downstream performance.

\begin{figure*}
\centering
\includegraphics[width=0.95\textwidth]{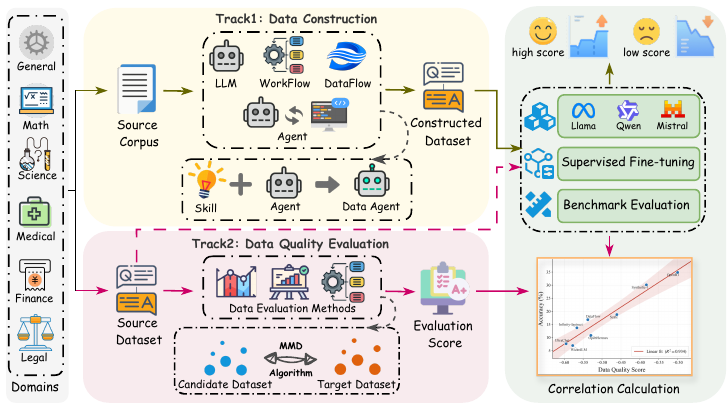}
\caption{The overall framework of DataPrep-Bench.}
\label{Frame}
\end{figure*}

We address this with \textbf{DataPrep-Bench}, a unified, downstream-grounded benchmark that evaluates both capabilities under the same raw sources, the same base models, the same training protocol, and the same downstream benchmarks (Figure~\ref{Frame}). The benchmark is organized as two tracks, each paired with a strong baseline:

\begin{itemize}
    \item \textbf{Data Construction Track.} Given raw domain sources, a construction method outputs a supervised dataset; we fine-tune a base model on it, jointly with Dolly-15k as a shared instruction-following corpus, and evaluate on held-out domain benchmarks. Our released baseline \textbf{Data-Construction-Skill} is a skill-guided agent that operationalizes construction through a reusable skill layer (output schemas, filtering rules, coverage constraints, and validation utilities) rather than a one-off prompt.

    \item \textbf{Data Quality Evaluation Track.} Given a pool of candidate training datasets, a scoring function assigns each a scalar, and we measure how well those scalars linearly predict the downstream performance of models fine-tuned on the same datasets; we release the ground-truth performance records alongside the candidate pools. Our released baseline \textbf{Distributional Alignment Score (DAS)} is a distribution-based evaluator that measures the Maximum Mean Discrepancy (MMD) between a candidate dataset and a domain proxy.
\end{itemize}

Our experiments support three takeaways.

\textbf{(T1) Adding synthesized domain data on top of Dolly-15k often hurts downstream performance}, across DataFlow-based, direct-LLM, and agent-based generators and both base models. Since the only difference between the Dolly-only baseline and the other rows is the domain-specific synthetic data, the regressions trace directly to that data, a conclusion that surface-level quality proxies would miss. This result is itself a call to action: the community needs stronger data construction methods, and DataPrep-Bench provides the end-to-end testbed under which such methods can be developed and fairly compared.

\textbf{(T2) No single family of construction methods is universally best.} DataFlow-Skill leads on structured domains (Finance, Law); agent-based methods lead in reasoning-heavy domains (Math, Medical); our Data-Construction-Skill is strongest in knowledge-extraction-dense domains, lifting the Dolly-only baseline by nearly 20 points absolute on Llama-3.1-8B Finance and matching the strongest agent- and DataFlow-based methods. Science and parts of Law remain open.

\textbf{(T3) DAS is the most reliable quality-evaluation metric overall.} It attains the strongest domain-averaged correlation with downstream performance in four of six domains, and is the only metric that simultaneously clears $r>0.70$ in Math, Science, and Medical. Existing quality- and diversity-based evaluators are either narrow specialists or sign-inconsistent across (domain, model) cells; Finance and Law remain difficult for every metric we tested.

Our contributions are threefold:

\begin{enumerate}
    \item We propose \textbf{DataPrep-Bench}, the first unified downstream-grounded benchmark that jointly covers LLM-driven data construction and data quality evaluation under shared domains, base models, training protocols, and downstream benchmarks.
    \item We release \textbf{Data-Construction-Skill}, a skill-guided agent baseline for the Data Construction Track, together with the curated raw source corpus.
    \item We release the \textbf{Distributional Alignment Score (DAS)}, a distribution-based baseline for the Data Quality Evaluation Track, together with the candidate pools and their ground-truth downstream performance records.
\end{enumerate}
\section{Background and Related Works}

\subsection{LLM-based Data Construction}

LLM-driven data construction has become a standard technique for building training and post-training corpora~\cite{JCST-2509-15948, bai2024survey}. Existing methods roughly fall into three groups.

The first group synthesizes instruction data from a small set of seeds. Self-Instruct~\cite{wang2023self} bootstraps instructions from human-written seeds, WizardLM~\cite{xu2304wizardlm} rewrites instructions to make them harder and more diverse, and UltraChat~\cite{UltraChat} scales multi-turn dialogues by sampling topics and roles. Tülu~3~\cite{lambert2024tulu3} and Infinity-Instruct~\cite{li2025infinityinstructscalinginstruction} extend this idea to large mixed-skill post-training corpora. These methods replace expensive human annotation with LLM generation, but the resulting data is anchored in the generator's own prior rather than in any external domain knowledge.

The second group grounds generation in long-form, authoritative domain materials such as books, papers, and technical documents, producing domain-specific SFT corpora like UltraMedical~\cite{zhang2024ultramedical} and MegaScience~\cite{fan2025megascience} (more representative datasets per domain are listed in Table~\ref{tab:domain-dataset-matrix}). This setting is the one our Data Construction Track targets: supervision has to be extracted and reformulated from curated domain sources rather than generated from scratch. The problem is that these works use very different sources, models, and training recipes, so cross-method comparison remains anecdotal.

The third group treats data construction as a system rather than a single prompt. Workflow-style frameworks such as DataFlow~\cite{liang2025dataflow}, Text2SQL-Flow~\cite{cai2025text2sql}, and step-level verification~\cite{shen2025let} split construction into reusable stages like chunking, filtering, scoring, rewriting, and verification. More recent efforts build fully agentic pipelines in which an LLM agent plans over a corpus, calls tools, and iteratively generates and validates samples. In practice, these pipelines still hard-code most of their control logic inside the agent's prompt, and there is no shared way to specify what counts as valid supervision at the corpus level. Our Data-Construction-Skill baseline addresses this by moving task semantics, output schemas, and validation rules into a reusable skill layer instead of hiding them inside a prompt.

\subsection{Data Quality Evaluation for LLM Training}

As candidate training data for LLMs keeps growing~\cite{llama, openai2023gpt, bai2024survey, JCST-2509-15948}, a natural question is how to tell in advance which data will actually help. Recent work shows that carefully curated high-quality data can match or beat much larger noisy corpora~\cite{zhou2023lima, gunasekar2023textbooks}. Existing quality-evaluation methods, which we survey below, differ mainly in the granularity at which they operate.

Most existing methods score each training example individually and then aggregate the scores. One line of work uses a strong model as a judge: QuRating~\cite{wettig2024qurating} learns pairwise quality preferences along four axes (Writing Style, Required Expertise, Facts \& Trivia, and Educational Value), each of which can be used as an independent scorer, and Liu et al.~\cite{liu2024what} systematically compare automatic quality scorers for alignment data and release Deita-Quality together with a separate Deita-Complexity scorer that measures instruction complexity as a diversity-oriented signal. Classifier-based scorers such as FineWeb-Edu,\footnote{\url{https://huggingface.co/HuggingFaceFW/fineweb-edu-classifier}} trained on Llama-3 labels of educational value, and PairQual,\footnote{\url{https://huggingface.co/zks2856/PairQual-Scorer-en}} trained on pairwise quality preferences, give cheap per-sample estimates; reward-model-style scorers apply a preference-tuned LLM as a proxy quality head on each sample. Model-side signals provide an alternative route: Superfiltering~\cite{li-etal-2024-superfiltering} uses loss differences between a weak and a strong model, LESS~\cite{xia2024less} uses gradient-based influence to select samples aligned with a target capability, and sample-level perplexity computed with a fixed reference model is another widely used lightweight indicator. These methods work well for per-sample filtering, but once we care about the quality of a dataset as a whole, they collapse to heuristics like averaging or top-$k$ selection, and are typically validated on a single backbone and a single domain.

A smaller line of work looks at the dataset as a whole rather than as a bag of samples. Diversity-based methods try to capture how much variation a dataset covers. The Vendi Score~\cite{friedman2022vendi} applies a similarity-matrix eigenvalue-based diversity estimator, and can be instantiated with different sentence encoders (e.g., BERT and SimCSE in our experiments) to yield BERTVendi and SimCSEVendi variants. Lexical-diversity measures~\cite{ploeger2024towards}, including MTLD and HD-D, quantify surface-form vocabulary variety; simple $n$-gram distinctness provides a related cheap baseline. Embedding-based methods such as Task2Vec~\cite{achille2019task2vec} instead summarize a dataset via a single ``task embedding'' derived from Fisher information, giving a compact proxy for dataset identity and difficulty. Distribution-alignment methods go one step further and compare the feature distribution of a candidate dataset against a reference that reflects the target capability; classical domain-adaptation results~\cite{redko2020survey} show that the discrepancy between training and target distributions directly bounds the gap between source and target risk. Our DAS baseline takes this view and measures quality as the Maximum Mean Discrepancy between a candidate dataset and a domain proxy.

\subsection{Benchmarks for LLM Data Preparation}
Several data-centric benchmarks evaluate data preparation strategies through downstream task performance. DataComp~\cite{gadre2023datacomp} benchmarks filtering strategies over image-text pairs via CLIP performance; its language-model counterpart DataComp-LM~\cite{li2024datacomp} extends this to text corpora, evaluating curation pipelines by the downstream performance of LLMs trained on the curated data. DataPerf~\cite{mazumder2023dataperf} provides leaderboards for data selection, debugging, and cleaning across modalities, while DCBENCH~\cite{cui2022dc} targets tabular tasks such as feature engineering and slice discovery. However, these benchmarks focus on traditional curation strategies (e.g., heuristic filtering, deduplication) rather than LLMs as autonomous data preparators. Meanwhile, data processing systems like Data-Juicer~\cite{chen2024datajuicer} and DataFlow~\cite{liang2025dataflow} offer modular tooling but do not benchmark the comparative effectiveness of different approaches. DataPrep-Bench fills this gap as the first benchmark that directly evaluates how well LLMs perform data preparation, spanning both data construction and data quality evaluation, grounded in end-to-end downstream performance across multiple domains and model architectures.

\section{Benchmark Design}

DataPrep-Bench evaluates two stages of LLM-driven data preparation. The \textbf{Data Construction Track} starts from raw domain source corpora and asks whether a method can turn them into useful supervised fine-tuning (SFT) data. The \textbf{Data Quality Evaluation Track} starts from a pool of candidate training datasets and asks whether a scoring function can predict which ones will improve downstream models, before any training is run. The two tracks take different inputs, but they share one principle: a data preparation decision counts only if it leads to better models.

Let the domain collection be $\mathcal{D} = \{D_1, D_2, \dots, D_m\}$. For each domain $D_k$, let $\mathcal{T}_k$ denote the corresponding downstream benchmark set and $f_0$ a base model. For any candidate SFT dataset $\mathcal{X}$ associated with domain $D_k$, we define
\begin{equation}
f(\mathcal{X}) = \mathrm{SFT}(f_0, \mathcal{X})
\end{equation}
as the adapted model obtained by fine-tuning $f_0$ on $\mathcal{X}$. Let $\mathrm{Perf}(f, \mathcal{T}_k)$ denote the downstream evaluation score of model $f$ on benchmark set $\mathcal{T}_k$.

\subsection{Data Construction Track}
\label{sec:benchmark_data_synthesis}

The Data Construction Track evaluates whether a method can transform raw domain knowledge sources into useful SFT data. In the current benchmark, the source corpus for each domain consists of a curated collection of domain-specific books and long-form knowledge materials. We consider the task of automatically constructing question-answer (QA) supervision data from these sources and using the resulting data for supervised fine-tuning. For domain $D_k$, let the raw source collection be $\mathcal{B}_k = \{b_{k,1}, b_{k,2}, \dots, b_{k,B_k}\}$, where each $b_{k,j}$ is a representative source document covering core content in $D_k$. Given a data construction method $M$, the goal is to transform $\mathcal{B}_k$ into an SFT dataset:
\begin{equation}
\mathcal{X}_k^{(M)} = M(\mathcal{B}_k)
= \{(q_i, a_i)\}_{i=1}^{N_k^{(M)}},
\end{equation}
where $q_i$ and $a_i$ denote the generated question and answer, respectively, and $N_k^{(M)}$ is the resulting dataset size. Different methods may produce datasets that differ substantially in scale, coverage, style, and quality.

The downstream performance of method $M$ in domain $D_k$ is defined as
\begin{equation}
\mathrm{Score}_{\mathrm{con}}(M, D_k)
=
\mathrm{Perf}\bigl(f(\mathcal{X}_k^{(M)}), \mathcal{T}_k\bigr).
\end{equation}
That is, we evaluate a construction method by the benchmark performance induced by its synthesized dataset. When $\mathcal{T}_k$ captures the core capabilities of the domain and is curated independently of the source corpus, this score serves as an operational proxy for the practical training utility of the constructed dataset, rather than merely its surface-level quality.

\subsection{Source Corpus for Data Construction}
\label{sec:source_corpus}

The Data Construction Track requires raw, non-trainable domain sources as input. The source corpus consists primarily of domain-specific books and long-form knowledge materials, intentionally chosen to differ from typical SFT corpora in three ways: (i)~the content is authored by domain experts rather than model-generated, (ii)~supervision must be \emph{extracted and reformulated} from these sources rather than directly consumed, and (iii)~each source is long (typically hundreds of pages) so that non-trivial chunking, coverage tracking, and validation are required. These properties make the corpus a realistic test bed for corpus-level data construction methods.

\paragraph{Collection procedure.}
For the Science, Law, Finance, and Medical domains, we collect books from three open textbook repositories: Wikibooks,\footnote{\url{https://www.wikibooks.org}} FreeBookCentre,\footnote{\url{https://www.freebookcentre.net}} and Open Textbook Library,\footnote{\url{https://open.umn.edu/opentextbooks}} which together span a wide range of disciplines. Within each domain we hand-select representative books to ensure coverage of major subfields, targeting at least three books per subfield, with each book typically containing hundreds of pages. For Math, we additionally collect Chinese university mathematics textbooks and Chinese mathematics competition materials, ensuring coverage of both foundational and olympiad-level problem styles. For Science, beyond the above textbook sources, we include preparatory and problem materials from the international science olympiads IPhO (Physics), IChO (Chemistry), and IBO (Biology) to inject high-difficulty reasoning content. For General Text, we instead use the FineWeb dataset\footnote{\url{https://huggingface.co/datasets/HuggingFaceFW/fineweb}} (a cleaned Common-Crawl derivative) and randomly sample 150\,MB of text from its sample-10BT subset, organized into 100 Markdown files, so that General Text reflects a web-scale distribution rather than curated books.

\paragraph{Unified input format.}
To give all construction methods the same kind of input, every PDF source is converted to Markdown using \texttt{MinerU}\footnote{\url{https://github.com/opendatalab/MinerU}}. After conversion, each domain $D_k$ exposes a folder of Markdown files that preserves section structure, tables, and formulas as faithfully as MinerU allows. Methods then differ only in how they process this common input (from single-pass prompting, through expert-designed workflows, to agentic pipelines), which is what allows their downstream scores to be compared end-to-end. The same Markdown folders are released as part of the benchmark artifact, so that future data construction methods can be evaluated under identical input conditions.

\subsection{Data Quality Evaluation Track}
\label{sec:benchmark_data_quality}

\paragraph{Motivation.}
Most existing data-quality metrics are either released as standalone scorers, with correlation studies on one or two datasets in a single domain, or validated indirectly through top-$k$ filtering, where a metric is deemed ``good'' if training on its top-ranked subset yields a gain. Both setups leave the central question unanswered: if we had to score candidate training datasets \emph{before paying the cost of fine-tuning}, how reliable would each metric be? The Data Quality Evaluation Track isolates exactly this question and turns it into a measurable property. Rather than comparing metrics through downstream gains on hand-picked datasets, it asks whether a metric's scalar output linearly tracks the empirical downstream performance of models fine-tuned on those datasets, under a shared protocol.

\paragraph{Formulation.}
Given a data quality metric or scoring function $Q$, which takes a candidate dataset as input and outputs a scalar score, the track asks a single question: to what extent does $Q$ linearly predict the downstream performance $\mathrm{Perf}(f(\cdot), \mathcal{T}_k)$ defined above? Although we use the term ``data quality,'' the target notion of quality here is \emph{downstream training utility} rather than surface-level textual fluency, cleanliness, or diversity.

\paragraph{Inputs and outputs.}
For each domain $D_k$, the track is built on a curated candidate dataset pool $\mathbb{X}_k = \{\mathcal{X}_{k,1}, \dots, \mathcal{X}_{k,C_k}\}$, where each $\mathcal{X}_{k,i}$ is a candidate training dataset sampled from widely used public SFT corpora. The pool covers both in-domain and out-of-domain distributions relative to $D_k$, allowing the benchmark to test whether a metric can distinguish datasets with different degrees of target-domain utility.

The track takes as input:
\begin{itemize}
    \item the candidate dataset pool $\mathbb{X}_k$;
    \item a target benchmark set $\mathcal{T}_k$ that captures the downstream capability of interest;
    \item the base model $f_0$ and the shared fine-tuning protocol used to obtain downstream performance; and
    \item optionally, a domain proxy dataset used by metrics that require an observable approximation of the target-domain distribution.
\end{itemize}
The output of the track, for a given metric $Q$, is a per-domain predictability score that quantifies how well the metric scores $Q(\mathcal{X}_{k,i})$ track the empirical downstream performance of models fine-tuned on the corresponding datasets.

\paragraph{Grounding in downstream performance.}
For each candidate dataset $\mathcal{X}_{k,i} \in \mathbb{X}_k$, the protocol produces two paired quantities:
\begin{equation}
q_{k,i} = Q(\mathcal{X}_{k,i}),
\qquad
s_{k,i} = \mathrm{Perf}\bigl(f(\mathcal{X}_{k,i}), \mathcal{T}_k\bigr).
\end{equation}
The score $q_{k,i}$ is a metric prediction computed before downstream fine-tuning, while $s_{k,i}$ is the empirical ground truth obtained by fine-tuning $f_0$ on $\mathcal{X}_{k,i}$ and evaluating the adapted model on $\mathcal{T}_k$. This reflects the intended use case of data quality evaluation: ranking or selecting candidate datasets before incurring the cost of downstream training. A metric is considered useful to the extent that its prediction vector
$\mathbf{q}_k = (q_{k,1}, \dots, q_{k,C_k})$
tracks the linear trend of
$\mathbf{s}_k = (s_{k,1}, \dots, s_{k,C_k})$.

\paragraph{Linear predictability as the evaluation criterion.}
We assess the extent to which $Q$ linearly predicts downstream utility in domain $D_k$ via the Pearson correlation between $\mathbf{q}_k$ and $\mathbf{s}_k$:
\begin{equation}
\rho(Q, D_k)
=
\frac{\sum_{i=1}^{C_k} (q_{k,i} - \bar{q}_k)(s_{k,i} - \bar{s}_k)}
{\sqrt{\sum_{i=1}^{C_k} (q_{k,i} - \bar{q}_k)^2}
 \sqrt{\sum_{i=1}^{C_k} (s_{k,i} - \bar{s}_k)^2}},
\end{equation}
where $\bar{q}_k$ and $\bar{s}_k$ are the sample means of $\{q_{k,i}\}_{i=1}^{C_k}$ and $\{s_{k,i}\}_{i=1}^{C_k}$, respectively. We report $\rho(Q, D_k)$ together with its two-sided $p$-value, and interpret a metric as a reliable predictor only when the correlation is statistically significant ($p < 0.05$). Because the evaluation is instantiated for multiple base models $f_0$, we report per-model correlations and also report domain-level summaries obtained by averaging $\rho(Q, D_k)$ across base models. An aggregate cross-domain summary is further computed when all metrics share the same score direction (i.e., higher scores correspond to higher predicted utility), enabling a fair overall comparison. We choose Pearson over rank-based alternatives because the ground-truth downstream scores $s_{k,i}$ lie on a meaningful numerical scale (benchmark accuracy), so that a useful metric should capture not only the ordering of candidates but also their relative spacing.

\paragraph{What the protocol accommodates.}
The formulation is deliberately agnostic to the mechanism of $Q$. Any scoring function can be plugged into the same pipeline and compared under identical candidate pools, base models, fine-tuning protocols, and downstream benchmarks, including: (i) \emph{per-sample quality scorers} such as QuRating, Deita-Quality, Superfiltering, FineWeb-Edu, PairQual, reward-model scorers, and perplexity, whose dataset-level value we obtain by averaging; (ii) \emph{dataset-level quality estimators} such as Vendi-Score variants (BERTVendi, SimCSEVendi); (iii) \emph{diversity measures} such as MTLD, HD-D, $n$-gram distinctness, Task2Vec, and Deita-Complexity; and (iv) \emph{distribution-alignment metrics} such as our proposed DAS, which compare the candidate distribution against a domain proxy. In Section~\ref{sec:methods}, we use this protocol to benchmark all of the above families head-to-head.

\subsection{Candidate and Proxy Datasets for Data Quality Evaluation}
\label{sec:candidate_proxy}

The Data Quality Evaluation Track requires two complementary resources per domain: a \emph{candidate pool} whose elements can be ranked by any metric $Q$, and a \emph{proxy dataset} used by metrics that compare a candidate against an observable approximation of the target-domain distribution. The concrete instantiation used in our benchmark is summarized in Table~\ref{tab:domain-dataset-matrix}; the design principles behind that instantiation are as follows.

\paragraph{Candidate pool design.}
For each domain $D_k$, the candidate pool $\mathbb{X}_k$ is instantiated with public SFT corpora spanning both in-domain and out-of-domain distributions relative to $D_k$. We intentionally mix the two: an all-in-domain pool would compress downstream scores into a narrow range and make correlation estimates unstable, while an all-out-of-domain pool would trivially be solved by any metric that detects ``this candidate is clearly off-topic.'' Mixing in-domain and out-of-domain candidates instead creates a spread of ground-truth downstream utilities that a metric must linearly track. Within this design, the pool size per domain varies with the availability of publicly released, non-contaminating in-domain SFT corpora: General Text contains the largest pool (14 candidates), while reasoning-intensive and specialized domains (Math, Science, Medical, Finance, Law) range from 8 to 10 candidates. This asymmetry is itself a benchmark signal, since it reflects the current scarcity of high-quality in-domain SFT data outside General Text, and it also interacts with the statistical power of the Pearson correlation: smaller pools require larger effect sizes to reach $p<0.05$ (cf.~Section~\ref{sec:results_quality}).

\paragraph{Proxy dataset selection.}
For metrics that require a reference distribution (notably DAS), we additionally select a high-quality in-domain proxy dataset per domain. The proxy plays the role of an observable anchor for the otherwise unavailable target-domain distribution $P_k^\star$. We select proxies according to three criteria: (i) \emph{domain alignment}: the proxy should represent the core capability of $D_k$ (e.g., mathematical reasoning for Math, clinical reasoning for Medical); (ii) \emph{data scale and quality}: it should be a well-established public SFT corpus with curated data, so that the proxy feature distribution is a faithful sample of the target distribution; and (iii) \emph{disjointness from the candidate pool and the downstream test set}, to prevent a metric from trivially scoring a candidate high simply because it overlaps with the proxy, and to avoid benchmark-set leakage. The proxy is used only for metric computation and is not trained on in either track. The concrete proxies used in our experiments (Infinity-Instruct for General, ODA-Math-460k for Math, Logics-STEM for Science, ReasonMed for Medical, Fin-o1 for Finance, and DISC-Law-SFT for Law) are listed in the top row of Table~\ref{tab:domain-dataset-matrix}.

\paragraph{What the candidate pool is not.}
Two points are worth emphasizing. First, the candidate pool is \emph{not} the benchmark training set: the downstream training utility $s_{k,i}$ of each candidate is computed by actually fine-tuning $f_0$ on $\mathcal{X}_{k,i}$, and this empirical training utility, not any notion of intrinsic dataset quality, is the ground truth that metrics must predict. Second, the pool is \emph{not} held fixed in size across domains: the benchmark is designed so that new in-domain candidates can be added to a given $\mathbb{X}_k$ without invalidating previously computed metric scores, since every metric operates on each candidate independently and the Pearson correlation is recomputed from scratch. We therefore view the current candidate pool as an initial release, and expect the pool sizes for specialized domains (Finance, Law, Medical) to grow as more high-quality in-domain SFT corpora become public.

\newcommand{\ds}[1]{\begin{tabular}[t]{@{}l@{}}#1\end{tabular}}
\begin{table*}[t]
\centering
\caption{Proxy datasets and candidate dataset pools across six domains.}
\label{tab:domain-dataset-matrix}
\setlength{\tabcolsep}{4pt}
\renewcommand{\arraystretch}{1.12}

\resizebox{\textwidth}{!}{%
\begin{tabular}{@{}lllllll@{}}
\toprule
\textbf{Category}
& \multicolumn{1}{c}{\textbf{General}}
& \multicolumn{1}{c}{\textbf{Math}}
& \multicolumn{1}{c}{\textbf{Science}}
& \multicolumn{1}{c}{\textbf{Medical}}
& \multicolumn{1}{c}{\textbf{Finance}}
& \multicolumn{1}{c}{\textbf{Law}} \\
\midrule

\ds{\textbf{Proxy /}\\ \textbf{reference}\\ \textbf{dataset}}
& \ds{Infinity-Instruct~\cite{li2025infinityinstructscalinginstruction}}
& \ds{ODA-Math-460k~\cite{oda-math}}
& \ds{Logics-STEM~\cite{xu2026logicsstemempoweringllmreasoning}}
& \ds{ReasonMed~\cite{sun2025reasonmed370kmultiagentgenerated}}
& \ds{Fin-o1~\cite{qian2025fino1}}
& \ds{DISC-Law-SFT~\cite{yue2023disclawllm}} \\
\midrule

\ds{\textbf{In-domain}\\ \textbf{candidate}\\ \textbf{dataset}}
& \ds{
DataFlow-10k~\cite{liang2025dataflow} \\
WizardLM 196k~\cite{xu2304wizardlm} \\
UltraChat~\cite{UltraChat} \\
Tulu 3 SFT Mixture~\cite{lambert2024tulu3} \\
Smoltalk Chinese~\cite{Smoltalk-Chinese} \\
OpenHermes 2.5$^{e}$
}
& \ds{
OpenR1-Math-220k$^{a}$ \\
Synthetic-1$^{b}$ \\
ScaleQuest~\cite{ding2024unleashing}
}
& \ds{
MegaScience~\cite{fan2025megascience} \\
Nemotron-Science-v1$^{c}$
}
& \ds{
UltraMedical~\cite{zhang2024ultramedical}
}
& \ds{
Finance-Instruct-500k$^{d}$
}
& \ds{
Lawyer-Llama~\cite{huang2023lawyer}
} \\
\midrule

\ds{\textbf{Out-domain}\\ \textbf{candidate}\\ \textbf{dataset}}
& \ds{
OpenR1-Math-220k$^{a}$ \\
Synthetic-1$^{b}$ \\
ScaleQuest~\cite{ding2024unleashing} \\
UltraMedical~\cite{zhang2024ultramedical} \\
Finance-Instruct-500k$^{d}$ \\
MegaScience~\cite{fan2025megascience} \\
Nemotron-Science-v1$^{c}$ \\
Lawyer-Llama~\cite{huang2023lawyer}
}
& \ds{
Infinity-Instruct~\cite{li2025infinityinstructscalinginstruction} \\
DataFlow-10k~\cite{liang2025dataflow} \\
WizardLM 196k~\cite{xu2304wizardlm} \\
UltraChat~\cite{UltraChat} \\
Tulu 3 SFT Mixture~\cite{lambert2024tulu3} \\
Smoltalk Chinese~\cite{Smoltalk-Chinese} \\
OpenHermes 2.5$^{e}$
}
& \ds{
Infinity-Instruct~\cite{li2025infinityinstructscalinginstruction} \\
DataFlow-10k~\cite{liang2025dataflow} \\
WizardLM 196k~\cite{xu2304wizardlm} \\
UltraChat~\cite{UltraChat} \\
Tulu 3 SFT Mixture~\cite{lambert2024tulu3} \\
Smoltalk Chinese~\cite{Smoltalk-Chinese} \\
OpenHermes 2.5$^{e}$
}
& \ds{
Infinity-Instruct~\cite{li2025infinityinstructscalinginstruction} \\
DataFlow-10k~\cite{liang2025dataflow} \\
WizardLM 196k~\cite{xu2304wizardlm} \\
UltraChat~\cite{UltraChat} \\
Tulu 3 SFT Mixture~\cite{lambert2024tulu3} \\
Smoltalk Chinese~\cite{Smoltalk-Chinese} \\
OpenHermes 2.5$^{e}$
}
& \ds{
Infinity-Instruct~\cite{li2025infinityinstructscalinginstruction} \\
DataFlow-10k~\cite{liang2025dataflow} \\
WizardLM 196k~\cite{xu2304wizardlm} \\
UltraChat~\cite{UltraChat} \\
Tulu 3 SFT Mixture~\cite{lambert2024tulu3} \\
Smoltalk Chinese~\cite{Smoltalk-Chinese} \\
OpenHermes 2.5$^{e}$
}
& \ds{
Infinity-Instruct~\cite{li2025infinityinstructscalinginstruction} \\
DataFlow-10k~\cite{liang2025dataflow} \\
WizardLM 196k~\cite{xu2304wizardlm} \\
UltraChat~\cite{UltraChat} \\
Tulu 3 SFT Mixture~\cite{lambert2024tulu3} \\
Smoltalk Chinese~\cite{Smoltalk-Chinese} \\
OpenHermes 2.5$^{e}$
} \\
\bottomrule
\end{tabular}%
}
\end{table*}
{\renewcommand\thefootnote{\alph{footnote}}%
\footnotetext[1]{\url{https://huggingface.co/datasets/open-r1/OpenR1-Math-220k}}%
\footnotetext[2]{\url{https://www.primeintellect.ai/blog/synthetic-1-release}}%
\footnotetext[3]{\url{https://huggingface.co/datasets/nvidia/Nemotron-Science-v1}}%
\footnotetext[4]{\url{https://huggingface.co/datasets/Josephgflowers/Finance-Instruct-500k}}%
\footnotetext[5]{\url{https://huggingface.co/datasets/teknium/OpenHermes-2.5}}}

\subsection{Downstream Evaluation Benchmarks}
\label{sec:benchmarks}

The target benchmark set $\mathcal{T}_k$ for each domain $D_k$, used to compute $\mathrm{Perf}(f, \mathcal{T}_k)$ in both tracks, comprises representative downstream tasks:
(1)~\textbf{General Text}: MMLU-Redux;
(2)~\textbf{Math}: AIME 2024, AMC 2023, Gaokao 2024, GSM8K, MATH, MinervaMath, and OlympiadBench, evaluated under the Qwen2.5-Math framework~\cite{yang2024qwen2};
(3)~\textbf{Science}: MMLU-STEM, MMLU-Pro, GPQA, SuperGPQA, ChemBench, PIQA, and SciBench, evaluated under the MegaScience~\cite{fan2025megascience} protocol;
(4)~\textbf{Medical}: MedR-Bench, MedMCQA, and MedCaseReasoning;
(5)~\textbf{Finance}: XFinBench, FinEval-KR, and CPA-KQA;
(6)~\textbf{Law}: LegalBench and LexGLUE.
For benchmarks with multiple subtasks, we report the average score. For tasks evaluated by exact-answer matching, we additionally apply an LLM-as-a-judge protocol uniformly across methods to account for semantically correct but surface-form-different answers. Per-benchmark descriptions are in Appendix~\ref{app:benchmarks}.
\section{Methods}
\label{sec:methods}

In this section, we present two concrete methodological instantiations used as strong baselines in DataPrep-Bench. For the Data Construction Track, we introduce \textbf{Data-Construction-Skill}, a skill-guided agentic method for transforming long-form domain sources into QA-style supervised fine-tuning data. For the Data Quality Evaluation Track, we introduce the \textbf{Distributional Alignment Score}, a distribution-based metric that estimates the downstream utility of a candidate dataset by measuring its alignment with a domain proxy.

\subsection{Data-Construction-Skill: Skill-Guided Agentic Data Construction}
\label{sec:data_construction_skill}

\subsubsection{Motivation and Overview}

Data-Construction-Skill is a skill-guided agentic method that turns long-form domain documents (e.g., textbooks, technical manuals) into reusable QA-style supervision data for post-training. The method is designed for a specific bottleneck: when supervision has to be extracted and reformulated from hundreds of pages of expert-authored content, no single prompt can reliably handle task decomposition, output consistency, validation, coverage, and resumable execution at once, and these limitations become especially pronounced when the generated data must remain faithful, diverse, and reusable across many sections of a long-form domain book. Our approach keeps the agent responsible for planning and execution but moves the \emph{specification of what counts as valid supervision} into a reusable skill layer. Formally, for each domain $D_k$, let $\mathcal{B}_k$ denote the source book corpus; Data-Construction-Skill, denoted $M_{\mathrm{DCS}}$, constructs a supervision dataset $\mathcal{X}_k^{(M_{\mathrm{DCS}})}$ from $\mathcal{B}_k$ through this skill-guided procedure.

A skill provides a reusable control layer that packages task instructions, output schemas, quality constraints, filtering rules, and auxiliary resources into a structured interface for agent execution. This makes it well suited to multi-step data construction over long-form corpora. Accordingly, our contribution is not merely a prompt for QA generation, but a skill-guided framework for agentic data construction. The skill specifies what constitutes valid supervision and how the task should be operationalized, while the agent remains responsible for planning over the corpus, processing chunks iteratively, and invoking supporting utilities during execution. This design preserves the flexibility of agentic planning without reducing the method to a rigid hand-written pipeline.

\subsubsection{Method Realization}

Given a domain corpus $\mathcal{B}_k$, we first partition it into a set of semantically coherent chunks $\mathcal{C}_k = \mathrm{Chunk}(\mathcal{B}_k)$, where $\mathrm{Chunk}(\cdot)$ denotes a structure-aware chunking procedure. The purpose of this step is to preserve local semantic completeness while providing the agent with manageable units for downstream construction.

For each chunk $c \in \mathcal{C}_k$, the agent first determines whether it contains reusable domain knowledge. Chunks that are primarily navigational, noisy, repetitive, or otherwise uninformative are discarded. For retained chunks, the agent identifies the main knowledge propositions and their local relations, including definitions, rules, mechanisms, conditions, exceptions, comparisons, and causal links. Based on this content, the method constructs up to three types of QA supervision for $c$, denoted $\mathcal{G}^{\mathrm{concept}}(c)$, $\mathcal{G}^{\mathrm{reason}}(c)$, and $\mathcal{G}^{\mathrm{case}}(c)$, each a finite set of QA pairs. These three sets are pairwise disjoint by construction, and we collect their union as the chunk-level supervision set:
\begin{equation}
\mathcal{G}(c)
=
\mathcal{G}^{\mathrm{concept}}(c)
\,\cup\,
\mathcal{G}^{\mathrm{reason}}(c)
\,\cup\,
\mathcal{G}^{\mathrm{case}}(c),
\end{equation}
where $\mathcal{G}^{\mathrm{concept}}(c)$ denotes concept-oriented QA pairs, $\mathcal{G}^{\mathrm{reason}}(c)$ denotes reasoning-oriented QA pairs that capture concise source-grounded reasoning patterns, and $\mathcal{G}^{\mathrm{case}}(c)$ denotes case-based QA pairs that place source-grounded knowledge into simple application scenarios. All generated supervision is instantiated as QA-style instruction-response pairs, ensuring consistency with the candidate dataset definition in Section~\ref{sec:benchmark_data_synthesis}. The final dataset is constructed as
\begin{equation}
\mathcal{X}_k^{(M_{\mathrm{DCS}})}
=
\bigcup_{c \in \mathcal{C}_k} \mathcal{G}(c).
\end{equation}

The three QA forms play complementary roles. Concept-oriented QA captures reusable atomic knowledge such as definitions, categories, functions, rules, and constraints. Reasoning-oriented QA captures short reasoning patterns that can be directly grounded in the source text. Case-based QA places the same knowledge into simple structured scenarios, encouraging transfer beyond isolated factual recall. These QA forms are generated adaptively according to the content of each chunk rather than being enforced uniformly.

A key role of the skill is to regulate this construction process. Instead of hard-coding the entire workflow, the skill provides the agent with reusable specifications for sample types, output schema, filtering criteria, coverage requirements, and validation behavior. In particular, the skill instructs the agent to produce QA pairs that are faithful to the source, self-contained without requiring access to the original passage, diverse in question type and difficulty, and useful as supervised fine-tuning examples. It also specifies invalid patterns, such as document-relative questions, hallucinated answers, duplicated samples, and questions whose answers cannot be supported by the source. In this sense, the skill serves as an intermediate control interface between high-level construction goals and low-level agent execution.

To improve the usability of the resulting dataset, we further apply post-processing and validation. We remove document-relative phrasing and passage-dependent questions, such as explicit references to the source text, chapter, section, or surrounding paragraph, so that the resulting data reflects reusable domain supervision rather than book-comprehension behavior. We also filter malformed, duplicated, or weakly grounded samples, and maintain chunk-level processing records to support long-document execution, coverage checking, and resumability.

Data-Construction-Skill can be viewed as a skill-guided agentic framework for transforming long-form domain books into reusable supervision data. Its novelty lies in introducing skills as a reusable control layer for data construction: rather than relying on one-shot prompting or a fixed pipeline, the method guides and stabilizes agent behavior through explicit skill-level instructions, schemas, filtering rules, and validation utilities.

\subsection{Distributional Alignment Score for Data Quality Evaluation}
\label{sec:method_das}

Recall from Section~\ref{sec:benchmark_data_quality} that the Data Quality Evaluation Track takes a candidate training dataset $\mathcal{X}_{k,i}$ as input and requires a scalar score $Q(\mathcal{X}_{k,i})$ as output. This score should be predictive of the downstream performance
\begin{equation}
s_{k,i}
=
\mathrm{Perf}(f(\mathcal{X}_{k,i}), \mathcal{T}_k).
\end{equation}
In this section, we propose one such metric, the Distributional Alignment Score (DAS), which instantiates $Q$ as a distribution-level similarity between the candidate dataset and a domain proxy, measured via the Maximum Mean Discrepancy (MMD).

\subsubsection{Theoretical Motivation}
\label{sec:das_theory}

DAS is motivated by domain adaptation theory. Let $\mathcal{F}$ denote the hypothesis class induced by a given model family, and let $h \in \mathcal{F}$ be a hypothesis trained on a source distribution $\mathcal{S}_X$ and evaluated on a target distribution $\mathcal{T}_X$. Theorem~36 of~\citet{redko2020survey} provides the following generalization bound: given two samples of size $m$ drawn i.i.d.\ from $\mathcal{S}_X$ and $\mathcal{T}_X$, respectively, with probability at least $1-\delta$, for all $h \in \mathcal{F}$,
\begin{equation}
\label{eq:theorem36}
\mathrm{R}^{\ell_q}_{\mathcal{T}}(h)
\leq
\mathrm{R}^{\ell_q}_{\mathcal{S}}(h)
+
d_{\mathrm{MMD}}(\hat{\mathcal{S}}_X, \hat{\mathcal{T}}_X)
+
\underbrace{
\frac{2}{m}\left(
\mathbb{E}_{\mathbf{x} \sim \mathcal{S}_X}
\left[\sqrt{\mathrm{tr}(K_{\mathcal{S}})}\right]
+
\mathbb{E}_{\mathbf{x} \sim \mathcal{T}_X}
\left[\sqrt{\mathrm{tr}(K_{\mathcal{T}})}\right]
\right)
+
2\sqrt{\frac{\log(2/\delta)}{2m}}
}_{\text{concentration and complexity terms}}
+
\lambda,
\end{equation}
where $\mathrm{R}^{\ell_q}_{\mathcal{T}}(h)$ and $\mathrm{R}^{\ell_q}_{\mathcal{S}}(h)$ denote the target and source risks, $d_{\mathrm{MMD}}$ is the empirical Maximum Mean Discrepancy, $K_{\mathcal{S}}$ and $K_{\mathcal{T}}$ are the kernel matrices on samples from $\mathcal{S}_X$ and $\mathcal{T}_X$, and $\lambda$ is the combined error of the ideal joint hypothesis.

This bound suggests that, under a fixed model family and training protocol, distributional alignment between the training source and the target is a principled factor governing target-side performance: a smaller $d_{\mathrm{MMD}}$ yields a tighter upper bound on target risk. Although this result does not directly prove downstream gains for LLM post-training, it motivates using distributional alignment between a candidate training dataset and the target-domain distribution as a training-free proxy for downstream utility.

\subsubsection{Definition of MMD on Text Datasets}
\label{sec:das_mmd}

To turn this intuition into a computable metric over text datasets, we first map text samples into a fixed feature space and then compute MMD between the induced feature distributions.

\paragraph{From text datasets to feature distributions.}
Each candidate dataset $\mathcal{X}_{k,i} = \{x_{k,i}^{(j)}\}_{j=1}^{N}$ is a set of textual samples, such as instruction--response pairs. We encode every sample with a fixed text encoder $\phi: \mathcal{X} \to \mathbb{R}^d$, producing a feature set
\begin{equation}
\Phi(\mathcal{X}_{k,i})
=
\{\phi(x_{k,i}^{(j)})\}_{j=1}^{N}
\subset \mathbb{R}^d,
\end{equation}
which we treat as samples from an induced feature distribution $P_{k,i}$ on $\mathbb{R}^d$. For each domain $D_k$, we additionally maintain a domain proxy dataset $\mathcal{X}^{\mathrm{proxy}}_k$, formally defined in Section~\ref{sec:das_definition}, which acts as an observable anchor of the target distribution. Encoding it with the same $\phi$ yields samples from an analogous proxy feature distribution $P^{\mathrm{proxy}}_k$. The encoder $\phi$ is fixed across all candidates and the proxy within a given evaluation, so that MMD values are directly comparable.

\paragraph{MMD in a reproducing kernel Hilbert space.}
Given a characteristic kernel $\kappa: \mathbb{R}^d \times \mathbb{R}^d \to \mathbb{R}$ with associated RKHS $\mathcal{H}_\kappa$ and feature map $\psi$, the population MMD between two distributions $P$ and $Q$ on $\mathbb{R}^d$ is defined as the RKHS distance between their mean embeddings:
\begin{align}
\mathrm{MMD}^2(P, Q; \kappa)
&=
\left\lVert
\mathbb{E}_{x \sim P}[\psi(x)]
-
\mathbb{E}_{y \sim Q}[\psi(y)]
\right\rVert_{\mathcal{H}_\kappa}^2
\notag \\
&=
\mathbb{E}_{x,x'\sim P}[\kappa(x,x')]
+
\mathbb{E}_{y,y'\sim Q}[\kappa(y,y')]
-
2\,\mathbb{E}_{x\sim P, y\sim Q}[\kappa(x,y)].
\label{eq:mmd_population}
\end{align}
For two finite samples $A = \{a_i\}_{i=1}^{n}$ and $B = \{b_j\}_{j=1}^{m}$ drawn from $P$ and $Q$, we use the biased empirical estimator
\begin{equation}
\label{eq:mmd_empirical}
\hat{d}_{\mathrm{MMD}}^{2}(A, B; \kappa)
=
\frac{1}{n^2} \sum_{i,i'} \kappa(a_i, a_{i'})
+
\frac{1}{m^2} \sum_{j,j'} \kappa(b_j, b_{j'})
-
\frac{2}{nm} \sum_{i,j} \kappa(a_i, b_j).
\end{equation}
We write
\begin{equation}
d_{\mathrm{MMD}}(A, B)
=
\sqrt{\max(0,\; \hat{d}_{\mathrm{MMD}}^{2}(A, B; \kappa))}.
\end{equation}
In our implementation, we use a Gaussian RBF kernel
\begin{equation}
\kappa(x, y)
=
\exp\left(-\frac{\lVert x-y\rVert_2^2}{2\sigma^2}\right),
\end{equation}
with a fixed bandwidth $\sigma$ shared across datasets. Keeping $\sigma$ fixed ensures that MMD values remain comparable across candidate datasets within the same evaluation. The Gaussian kernel is characteristic, implying that the population MMD is zero if and only if the two feature distributions are identical.

\subsubsection{The DAS Metric}
\label{sec:das_definition}

A direct application of Eq.~\eqref{eq:theorem36} would require computing MMD between each candidate training dataset and the benchmark test set, which would contaminate the benchmark because test data would influence data selection. We therefore adopt a proxy-based formulation.

\paragraph{Proxy dataset.}
For each domain $D_k$, we designate a proxy dataset $\mathcal{X}^{\mathrm{proxy}}_k$ that (i) approximates the target capability distribution of $D_k$, (ii) is drawn from a public SFT corpus aligned with the domain, and (iii) has no sample-level overlap with the benchmark test set $\mathcal{T}_k$. The proxy acts as an observable anchor of the otherwise unavailable target-domain distribution.

\paragraph{DAS definition.}
Given a candidate dataset $\mathcal{X}_{k,i}$, we first compute its proxy distance under the MMD induced by the encoder $\phi$ and kernel $\kappa$:
\begin{equation}
\label{eq:das_distance}
M_{k,i}
=
d_{\mathrm{MMD}}\!\left(
\Phi(\mathcal{X}_{k,i}),
\Phi(\mathcal{X}^{\mathrm{proxy}}_k);
\kappa
\right),
\end{equation}
using the empirical estimator in Eq.~\eqref{eq:mmd_empirical}. The Distributional Alignment Score is then defined as the negative MMD distance:
\begin{equation}
\label{eq:das_score}
\mathrm{DAS}(\mathcal{X}_{k,i})
=
- M_{k,i}.
\end{equation}
Thus, higher DAS indicates stronger distributional alignment with the domain proxy and, under the motivation above, higher expected downstream utility. DAS is the concrete instantiation of $Q$ that we plug into the evaluation protocol of Section~\ref{sec:benchmark_data_quality}.

\paragraph{Validity of the proxy.}
The proxy-based formulation is motivated by the triangle inequality of the population MMD metric over the induced feature distributions:
\begin{equation}
\label{eq:mmd_triangle}
d_{\mathrm{MMD}}(P_{k,i}, P_k^\star)
\leq
d_{\mathrm{MMD}}(P_{k,i}, P^{\mathrm{proxy}}_k)
+
d_{\mathrm{MMD}}(P^{\mathrm{proxy}}_k, P_k^\star),
\end{equation}
where $P_{k,i}$ is the feature distribution induced by candidate dataset $\mathcal{X}_{k,i}$, $P^{\mathrm{proxy}}_k$ is the feature distribution induced by the domain proxy, and $P_k^\star$ denotes the latent target-domain feature distribution underlying the benchmark set $\mathcal{T}_k$. While Eq.~\eqref{eq:mmd_triangle} does not guarantee exact preservation of the ranking induced by the unknown distance to $P_k^\star$, it shows that a proxy close to $P_k^\star$ introduces only a bounded additive slack. Therefore, DAS can serve as a surrogate signal for target proximity while avoiding benchmark contamination. We validate this choice empirically: a good metric, plugged into the protocol of Section~\ref{sec:benchmark_data_quality}, should produce scores $\{\mathrm{DAS}(\mathcal{X}_{k,i})\}_i$ that are strongly correlated with the observed downstream performance $\{s_{k,i}\}_i$.

\paragraph{Computation pipeline.}
Computing DAS for a candidate dataset involves three steps:
\begin{enumerate}
    \item \textbf{Encode.} Apply the fixed text encoder $\phi$ to every sample in $\mathcal{X}_{k,i}$ and $\mathcal{X}^{\mathrm{proxy}}_k$ to obtain feature sets $\Phi(\mathcal{X}_{k,i})$ and $\Phi(\mathcal{X}^{\mathrm{proxy}}_k)$.
    \item \textbf{Measure.} Compute the empirical MMD between the two feature sets using Eq.~\eqref{eq:mmd_empirical} with a Gaussian RBF kernel.
    \item \textbf{Score.} Return $\mathrm{DAS}(\mathcal{X}_{k,i}) = -d_{\mathrm{MMD}}\bigl(\Phi(\mathcal{X}_{k,i}), \Phi(\mathcal{X}^{\mathrm{proxy}}_k); \kappa\bigr)$.
\end{enumerate}
This pipeline requires no downstream fine-tuning, depends only on the input candidate and the fixed domain proxy, and is directly pluggable into the Data Quality Evaluation protocol.

\paragraph{Instantiation across domains.}
For each evaluation domain, we select a domain-specific proxy dataset: Infinity-Instruct~\cite{li2025infinityinstructscalinginstruction} for general text, ODA-Math-460k~\cite{oda-math} for mathematical reasoning, Logics-STEM~\cite{xu2026logicsstemempoweringllmreasoning} for science knowledge and reasoning, ReasonMed~\cite{sun2025reasonmed370kmultiagentgenerated} for medical expertise, Fin-o1~\cite{qian2025fino1} for financial expertise, and DISC-Law-SFT~\cite{yue2023disclawllm} for legal expertise. We then curate a library of mainstream candidate training datasets drawn from both in-domain and out-of-domain public SFT corpora, compute $\mathrm{DAS}(\mathcal{X}_{k,i})$ for each candidate, and assess the metric by measuring the Pearson correlation between $\{\mathrm{DAS}(\mathcal{X}_{k,i})\}_{i=1}^{C_k}$ and the downstream performance values $\{s_{k,i}\}_{i=1}^{C_k}$ defined in Section~\ref{sec:benchmark_data_quality}. Full implementation details, including encoder choice, sample size, bandwidth, and candidate pools, are provided in Section~\ref{sec:experiments_detail} and Appendix~\ref{app:impl}.
\begin{table*}[!t]
\centering
\caption{Performance of Qwen2.5-7B on Math, General, and Finance benchmarks after fine-tuning on datasets synthesized by different generators. All rows are fine-tuned on Dolly-15k (a general instruction-following corpus) jointly with the domain-specific dataset produced by the named generator; $^\dag$~marks the Dolly-15k only reference baseline, fine-tuned on Dolly-15k alone without any domain-specific synthetic data. Benchmark abbreviations: MM = MinervaMath, OB = OlympiadBench, M500 = MATH-500, GK24 = Gaokao2024, MR = MMLU-Redux, CKQA = CPA-KQA, FEKR = FinEval-KR, XFB = XFinBench.}
\label{tab:synthetic_data_mgf_qwen}
\resizebox{\textwidth}{!}{
\begin{tabular}{l|cccccccc|cc|cccc}
\toprule
\multirow{2}{*}{\textbf{Training Data Generator}} 
& \multicolumn{8}{c|}{\textbf{Math}} 
& \multicolumn{2}{c|}{\textbf{General}} 
& \multicolumn{4}{c}{\textbf{Finance}} \\
\cmidrule(lr){2-9}\cmidrule(lr){10-11}\cmidrule(lr){12-15}
& GSM8K & AMC23 & AIME24 & MM & OB & M500 & GK24 & Avg 
& MR & Avg 
& CKQA & FEKR & XFB & Avg \\
\midrule
Dolly-15k only$^\dag$ & 69.9 & 17.5 & \underline{0.0} & 10.7 & 10.7 & \textbf{39.8} & 16.5 & 23.6 & 77.7 & 77.7 & \underline{57.6} & \underline{59.4} & 56.3 & \underline{57.8} \\
\midrule
\rowcolor[rgb]{.867, .922, .969}
\multicolumn{15}{c}{\textit{\textbf{DataFlow-based Generators}}} \\
\midrule
DataFlow & 56.5 & 7.5 & \underline{0.0} & 7.7 & 7.9 & 27.4 & 17.6 & 17.8 & \underline{77.9} & \underline{77.9} & 51.0 & 54.5 & 59.3 & 54.9 \\
DataFlow-Skill & 56.7 & 10.0 & \underline{0.0} & 8.8 & 6.2 & 22.8 & \textbf{28.6} & 19.0 & 76.5 & 76.5 & \textbf{60.0} & \textbf{65.4} & \textbf{68.9} & \textbf{64.8} \\
\midrule
\rowcolor[rgb]{.867, .922, .969}
\multicolumn{15}{c}{\textit{\textbf{LLM-based Generators}}} \\
\midrule
Claude Opus 4.6 & 68.8 & 12.5 & \underline{0.0} & 8.8 & 9.8 & 35.5 & 14.3 & 21.4 & \textbf{78.2} & \textbf{78.2} & 37.6 & 39.6 & 55.9 & 44.4 \\
Gemini 3.0 Pro & 72.1 & 20.0 & \underline{0.0} & 10.7 & 11.4 & 37.9 & 15.4 & 23.9 & 77.8 & 77.8 & 48.6 & 49.5 & \underline{58.4} & 52.2 \\
GPT-5.2 & 66.7 & 17.5 & \underline{0.0} & 7.7 & 11.1 & 32.7 & 14.3 & 21.4 & \underline{77.9} & \underline{77.9} & 47.1 & 43.6 & 53.6 & 48.1 \\
\midrule
\rowcolor[rgb]{.867, .922, .969}
\multicolumn{15}{c}{\textit{\textbf{Agent-based Generators}}} \\
\midrule
Qwen3.5-Plus & \textbf{72.7} & \underline{22.5} & \textbf{3.3} & \underline{11.0} & \underline{11.6} & \underline{38.7} & 16.5 & 25.2 & 77.6 & 77.6 & 48.1 & 49.5 & 55.6 & 51.1 \\
GLM-4.7 & 71.4 & \underline{22.5} & \textbf{3.3} & 9.6 & \underline{11.6} & 37.9 & 20.9 & \underline{25.3} & 77.3 & 77.3 & 51.4 & 55.4 & 53.1 & 53.3 \\
Claude Opus 4.6 & 69.4 & 10.0 & \underline{0.0} & \underline{11.0} & 8.9 & 33.8 & 19.8 & 21.8 & 75.9 & 75.9 & 32.4 & 39.6 & 53.8 & 41.9 \\
Gemini 3.0 Pro & 70.1 & 15.0 & \underline{0.0} & 8.8 & 10.8 & 35.7 & 20.9 & 23.0 & 77.7 & 77.7 & 53.3 & 52.5 & 55.4 & 53.7 \\
GPT-5.2 & 69.6 & \textbf{25.0} & \textbf{3.3} & 8.8 & 9.9 & 35.6 & \underline{26.4} & \textbf{25.5} & 77.7 & 77.7 & 39.1 & 39.6 & 51.0 & 43.2 \\
GPT-5.3-codex & 70.8 & 15.0 & \underline{0.0} & \underline{11.0} & \textbf{11.7} & 38.5 & 13.2 & 22.9 & 77.6 & 77.6 & 58.6 & 63.4 & 55.9 & 59.3 \\
\midrule
\textbf{Skill~(Claude Opus 4.6)} & \underline{72.6} & 17.5 & \textbf{3.3} & \textbf{14.0} & 11.1 & 36.8 & 13.2 & 24.1 & \textbf{78.2} & \textbf{78.2} & \underline{57.6} & 53.5 & 55.4 & 55.5 \\
\bottomrule
\end{tabular}
}
\end{table*}

\section{Experiments}
\label{sec:experiments_detail}

We evaluate DataPrep-Bench across the two tracks defined in Section~\ref{sec:benchmark_data_synthesis} and Section~\ref{sec:benchmark_data_quality}. In the \textbf{Data Construction Track}, we compare data construction methods by fine-tuning base models on synthesized datasets and measuring downstream performance; the raw source corpus (Section~\ref{sec:source_corpus}) and the downstream benchmarks (Section~\ref{sec:benchmarks}) are fixed across all methods. In the \textbf{Data Quality Evaluation Track}, we assess whether data quality metrics can predict the downstream utility of candidate training datasets before fine-tuning; the candidate pools and proxy datasets (Section~\ref{sec:candidate_proxy}) and the same downstream benchmarks serve as ground truth. All construction methods consume the same raw corpus, and all quality metrics are evaluated on the same (candidate, ground-truth downstream score) pairs, so that cross-method and cross-metric comparisons are made under identical conditions.

\subsection{Experimental Setup}
\label{sec:experimental_setup}

\subsubsection{Data Construction}

We compare data construction methods that all consume the same raw source corpus (Section~\ref{sec:source_corpus}); each method emits a synthesized SFT dataset, which is then used to fine-tune a base model whose downstream performance on $\mathcal{T}_k$ is reported.

\paragraph{DataFlow-based generators.}
DataFlow executes multi-step expert-designed workflows for data construction. It offers a dedicated pipeline for cleaning and generating knowledge bases from Markdown-format books. DataFlow-Skill selects or creates custom operators and composes them into an executable pipeline using GPT-4o; note that this is a DataFlow-internal variant distinct from our proposed Data-Construction-Skill (Section~\ref{sec:data_construction_skill}), which is an agent-based method.

\paragraph{Direct LLM-based generators.}
Models including Claude Opus 4.6, Gemini 3.0 Pro, and GPT-5.2 are prompted to generate QA pairs from source books, with inputs truncated to fit the context window when necessary. This baseline reflects a simple direct-generation strategy; more sophisticated processing is covered by workflow- and agent-based methods.

\paragraph{Agent-based generators.}
Agents operate under a ReAct-style format with Claude Code as the execution environment. Backbones include Qwen3.5-Plus, GLM-4.7, Claude Opus 4.6, Gemini 3.0 Pro, and GPT-5.2, each evaluated as a standard-prompted agent whose instructions emphasize data quality, coverage, diversity, and faithfulness: the agent is asked to generate QA pairs covering core concepts, details, examples, and case studies, with varied question types, difficulty levels, and topics, while keeping each pair self-contained and grounded in the source text. Additionally, we evaluate a skill-guided variant using Data-Construction-Skill (Section~\ref{sec:data_construction_skill}) with Claude Opus 4.6 as the backbone, which replaces the prompt-level specification above with a reusable skill layer that encodes task instructions, output schemas, quality constraints, and validation rules.

\paragraph{Implementation details.}
We fine-tune with LlamaFactory~\cite{zheng2024llamafactory} under a consistent set of hyperparameters on two base models, Qwen2.5-7B~\cite{qwen2.5} and Llama-3.1-8B~\cite{grattafiori2024llama}. All runs jointly fine-tune on Dolly-15k\footnote{\url{https://huggingface.co/datasets/databricks/databricks-dolly-15k}} as a general instruction-following corpus to isolate the contribution of domain-specific synthetic data; the Dolly-15k only baseline (marked $^\dag$ in the tables) uses Dolly-15k exclusively, with no domain-specific synthetic data. Full hyperparameters, evaluation protocols, and the dataset-size design rationale are provided in Appendix~\ref{app:impl}.

\begin{table*}[!t]
\centering
\caption{Performance of Llama-3.1-8B on Math, General, and Finance benchmarks after fine-tuning on datasets synthesized by different generators. All rows are fine-tuned on Dolly-15k (a general instruction-following corpus) jointly with the domain-specific dataset produced by the named generator; $^\dag$~marks the Dolly-15k only reference baseline, fine-tuned on Dolly-15k alone without any domain-specific synthetic data. Benchmark abbreviations: MM = MinervaMath, OB = OlympiadBench, M500 = MATH-500, GK24 = Gaokao2024, MR = MMLU-Redux, CKQA = CPA-KQA, FEKR = FinEval-KR, XFB = XFinBench.}
\label{tab:synthetic_data_mgf_llama}
\resizebox{\textwidth}{!}{
\begin{tabular}{l|cccccccc|cc|cccc}
\toprule
\multirow{2}{*}{\textbf{Training Data Generator}} 
& \multicolumn{8}{c|}{\textbf{Math}} 
& \multicolumn{2}{c|}{\textbf{General}} 
& \multicolumn{4}{c}{\textbf{Finance}} \\
\cmidrule(lr){2-9}\cmidrule(lr){10-11}\cmidrule(lr){12-15}
& GSM8K & AMC23 & AIME24 & MM & OB & M500 & GK24 & Avg 
& MR & Avg 
& CKQA & FEKR & XFB & Avg \\
\midrule
Dolly-15k only$^\dag$ & \textbf{33.9} & \textbf{10.0} & 0.0 & 6.6 & 3.7 & \textbf{13.6} & 14.3 & \textbf{11.7} & \textbf{67.1} & \textbf{67.1} & 12.6 & 12.4 & 20.3 & 15.1 \\
\midrule
\rowcolor[rgb]{.867, .922, .969}
\multicolumn{15}{c}{\textit{\textbf{DataFlow-based Generators}}} \\
\midrule
DataFlow & 8.4 & \textbf{10.0} & 0.0 & 4.8 & 2.5 & 6.1 & 8.8 & 5.8 & 51.1 & 51.1 & 27.1 & 26.7 & 40.5 & 31.4 \\
DataFlow-Skill & 8.7 & 2.5 & 0.0 & 4.4 & 3.3 & 7.2 & \underline{16.5} & 6.1 & 48.4 & 48.4 & \underline{34.8} & \textbf{31.7} & \underline{43.0} & \textbf{36.5} \\
\midrule
\rowcolor[rgb]{.867, .922, .969}
\multicolumn{15}{c}{\textit{\textbf{LLM-based Generators}}} \\
\midrule
Claude Opus 4.6 & 28.7 & 5.0 & 0.0 & 6.2 & 2.8 & 11.2 & 11.0 & 9.3 & \underline{52.9} & \underline{52.9} & 22.9 & 24.8 & 37.5 & 28.4 \\
Gemini 3.0 Pro & 30.7 & \underline{7.5} & 0.0 & \textbf{8.1} & 3.7 & 12.6 & 14.3 & \underline{11.0} & 50.8 & 50.8 & 23.8 & \underline{27.7} & 42.1 & 31.2 \\
GPT-5.2 & 26.8 & 2.5 & 0.0 & \underline{7.7} & \underline{4.1} & 10.2 & 13.2 & 9.2 & 47.4 & 47.4 & 25.2 & 24.8 & 32.2 & 27.4 \\
\midrule
\rowcolor[rgb]{.867, .922, .969}
\multicolumn{15}{c}{\textit{\textbf{Agent-based Generators}}} \\
\midrule
Qwen3.5-Plus & \underline{33.7} & 5.0 & 0.0 & 6.6 & \textbf{4.4} & 11.3 & 12.1 & 10.4 & 50.4 & 50.4 & 23.3 & 20.8 & 38.2 & 27.4 \\
GLM-4.7 & 31.8 & 2.5 & 0.0 & 5.9 & 3.1 & \underline{13.0} & \textbf{17.6} & 10.6 & 49.2 & 49.2 & 23.8 & 25.7 & \underline{43.0} & 30.8 \\
Claude Opus 4.6 & 16.4 & 5.0 & 0.0 & 6.6 & 2.5 & 7.7 & 8.8 & 6.7 & 47.5 & 47.5 & 23.3 & 25.7 & 37.7 & 28.9 \\
Gemini 3.0 Pro & 29.6 & \underline{7.5} & 0.0 & 6.2 & 3.6 & 11.5 & \textbf{17.6} & 10.9 & 47.9 & 47.9 & 25.2 & 21.8 & 40.9 & 29.3 \\
GPT-5.2 & 28.7 & 5.0 & 0.0 & 6.6 & 4.0 & 10.4 & 15.4 & 10.0 & 49.0 & 49.0 & 21.9 & 21.8 & \textbf{44.8} & 29.5 \\
GPT-5.3-codex & 29.9 & 2.5 & 0.0 & 5.9 & 2.8 & 12.2 & 14.3 & 9.7 & 51.3 & 51.3 & 23.3 & 25.7 & 41.8 & 30.3 \\
\midrule
\textbf{Skill~(Claude Opus 4.6)} & 16.5 & \underline{7.5} & 0.0 & 4.4 & 3.0 & 9.5 & 11.0 & 7.4 & 51.3 & 51.3 & \textbf{36.2} & \textbf{31.7} & 34.7 & \underline{34.2} \\
\bottomrule
\end{tabular}
}
\end{table*}

\subsubsection{Data Quality Evaluation}

\paragraph{Baselines.}
We adopt 17 dataset quality and diversity evaluation operators from the DataFlow~\cite{liang2025dataflow} framework as baselines for DAS, giving 18 metrics in total when DAS is included. Gradient-based influence methods such as LESS~\cite{xia2024less} are not included because they require per-gradient computation over the target evaluation set, a cost that is prohibitive at the scale of our multi-domain candidate pools and three base models, and because they presuppose a known target task, which contradicts the task-agnostic protocol of our benchmark. Quality-oriented evaluators include QuratingSampleEvaluator~\cite{wettig2024qurating} (with four sub-variants: Writing Style, Required Expertise, Facts \& Trivia, and Educational Value), Deita-QualitySampleEvaluator~\cite{liu2024what}, RM-SampleEvaluator, SuperfilteringSampleEvaluator~\cite{li-etal-2024-superfiltering}, FineWeb-EduSampleEvaluator,\footnote{\url{https://huggingface.co/HuggingFaceFW/fineweb-edu-classifier}} PairQualSampleEvaluator,\footnote{\url{https://huggingface.co/zks2856/PairQual-Scorer-en}} PerplexityEvaluator, and VendiDatasetEvaluator~\cite{friedman2022vendi} (with two embedding variants: BERT and SimCSE). Diversity-oriented evaluators include LexicalDiversitySampleEvaluator~\cite{ploeger2024towards} (with two variants: MTLD and HD-D), N-gramSampleEvaluator, Task2VecDatasetEvaluator~\cite{achille2019task2vec}, and Deita-ComplexitySampleEvaluator~\cite{liu2024what}. Sample-level evaluator scores are aggregated by mean; dataset-level evaluators use raw output values.

\paragraph{Implementation details.}
To test whether the correlation between distributional alignment and downstream performance is model-dependent or broadly applicable, we use three base models spanning distinct architectures: Qwen2.5-7B~\cite{qwen2.5}, Llama-3.1-8B~\cite{grattafiori2024llama}, and Mistral-7B-v0.3~\cite{jiang2023mistral}; Mistral is included here (but not in the Data Construction Track) specifically to stress-test whether metric rankings are robust across model families. For DAS, we encode each candidate dataset with Qwen3-Embedding-8B (4096-dim embeddings), randomly sample 5{,}000 instances per dataset, and compute MMD with a Gaussian RBF kernel ($\sigma=1.0$). Full details are in Appendix~\ref{app:impl}.

\subsection{Results of Data Construction}
\label{sec:results_construction}

Tables~\ref{tab:synthetic_data_mgf_qwen}--\ref{tab:synthetic_data_slm_llama} report results across six domains and two base models. The picture that emerges is more nuanced than a simple ranking of construction methods.

\paragraph{Synthetic domain data often hurts rather than helps.}
The most consistent pattern in our results is not that one method wins; it is that adding domain-specific synthesized data on top of Dolly-15k frequently makes things worse. On Llama-3.1-8B, the Dolly-only baseline already achieves reasonable scores in Math and Science; once domain data is mixed in, most generators drag performance down rather than up, and the regression is not confined to any single family of methods. The two experimental conditions share the same instruction-following corpus and differ only in the presence of domain-specific synthetic data, so the regressions trace squarely back to that data. The implication is straightforward: ``more synthetic data'' is not a safe default, and surface-level quality metrics that would score this data highly clearly miss something important. Downstream-grounded evaluation is not just a nice-to-have.

\begin{table*}[!t]
\centering
\caption{Performance of Qwen2.5-7B on Science, Law, and Medical benchmarks after fine-tuning on datasets synthesized by different generators. All rows are fine-tuned on Dolly-15k (a general instruction-following corpus) jointly with the domain-specific dataset produced by the named generator; $^\dag$~marks the Dolly-15k only reference baseline, fine-tuned on Dolly-15k alone without any domain-specific synthetic data. Benchmark abbreviations: LB = LegalBench, LG = LexGLUE, MCR = MedCaseReasoning, MMCQA = MedMCQA, MRB = MedR-Bench, MSTEM = MMLU-STEM, MPRO = MMLU-Pro, SGPQA = SuperGPQA, CB = ChemBench, SB = SciBench.}
\label{tab:synthetic_data_slm_qwen}
\resizebox{\textwidth}{!}{
\begin{tabular}{l|ccc|cccc|cccccccc}
\toprule
\multirow{2}{*}{\textbf{Training Data Generator}} 
& \multicolumn{3}{c|}{\textbf{Law}} 
& \multicolumn{4}{c|}{\textbf{Medical}} 
& \multicolumn{8}{c}{\textbf{Science}} \\
\cmidrule(lr){2-4}\cmidrule(lr){5-8}\cmidrule(lr){9-16}
& LB & LG & Avg 
& MCR & MMCQA & MRB & Avg 
& MSTEM & MPRO & GPQA & SGPQA & CB & PIQA & SB & Avg \\
\midrule
Dolly-15k only$^\dag$ & 86.9 & 62.0 & 74.5 & 13.6 & 27.4 & 67.8 & 36.3 & \textbf{47.5} & \textbf{28.6} & 21.1 & \textbf{15.7} & \textbf{24.3} & \textbf{52.7} & \underline{5.2} & \textbf{27.9} \\
\midrule
\rowcolor[rgb]{.867, .922, .969}
\multicolumn{16}{c}{\textit{\textbf{DataFlow-based Generators}}} \\
\midrule
DataFlow & 89.7 & \textbf{64.8} & \textbf{77.2} & 9.9 & 29.4 & 63.6 & 34.3 & 37.9 & 23.9 & 20.7 & 11.8 & 15.7 & 28.8 & 2.0 & 20.1 \\
DataFlow-Skill & \textbf{92.0} & 57.4 & 74.7 & 11.9 & 24.1 & 65.6 & 33.9 & 44.2 & 24.6 & 18.0 & 13.0 & 18.8 & 43.5 & 2.0 & 23.4 \\
\midrule
\rowcolor[rgb]{.867, .922, .969}
\multicolumn{16}{c}{\textit{\textbf{LLM-based Generators}}} \\
\midrule
Claude Opus 4.6 & 88.0 & 63.2 & \underline{75.6} & 13.9 & 8.1 & 66.8 & 29.6 & 38.7 & 24.6 & \underline{22.4} & 13.1 & 19.3 & 40.9 & 2.7 & 23.1 \\
Gemini 3.0 Pro & 85.9 & \underline{63.5} & 74.7 & 12.2 & 6.0 & \underline{69.1} & 29.1 & 39.7 & 25.7 & 19.7 & 12.7 & 21.2 & 44.0 & 3.3 & 23.8 \\
GPT-5.2 & 89.2 & 60.9 & 75.0 & 10.6 & 23.3 & 66.0 & 33.3 & 38.3 & 24.6 & \textbf{22.9} & 12.8 & 17.9 & 38.1 & 2.0 & 22.4 \\
\midrule
\rowcolor[rgb]{.867, .922, .969}
\multicolumn{16}{c}{\textit{\textbf{Agent-based Generators}}} \\
\midrule
Qwen3.5-Plus & \underline{90.2} & 61.0 & \underline{75.6} & \underline{16.5} & 16.5 & 68.3 & 33.8 & \textbf{47.5} & 26.9 & 21.7 & \underline{14.8} & \underline{22.9} & 48.6 & \textbf{5.3} & \underline{26.8} \\
GLM-4.7 & 84.8 & 61.4 & 73.1 & 15.2 & 10.6 & 66.8 & 30.9 & 42.0 & 25.7 & 18.9 & 12.7 & 19.4 & 43.6 & 2.3 & 23.5 \\
Claude Opus 4.6 & 65.2 & 48.8 & 57.0 & \textbf{16.6} & \underline{50.7} & 54.3 & \underline{40.5} & 36.6 & 24.1 & 19.0 & 12.9 & 19.2 & 44.1 & 2.2 & 22.6 \\
Gemini 3.0 Pro & 57.5 & 55.5 & 56.5 & 15.8 & \textbf{52.3} & 63.4 & \textbf{43.8} & \underline{46.0} & \underline{27.6} & \underline{22.4} & 14.7 & 20.3 & \underline{50.0} & 2.5 & 26.2 \\
GPT-5.2 & 61.7 & 51.9 & 56.8 & 14.7 & 45.0 & 59.0 & 39.6 & 44.0 & 25.3 & 21.1 & 14.0 & 20.2 & 46.4 & 2.8 & 24.8 \\
GPT-5.3-codex & 87.8 & 63.4 & \underline{75.6} & 13.6 & 20.3 & \textbf{70.0} & 34.6 & 35.9 & 22.4 & 16.4 & 11.2 & 16.5 & 33.1 & 3.4 & 19.8 \\
\midrule
\textbf{Skill~(Claude Opus 4.6)} & 85.5 & 61.7 & 73.6 & 9.9 & 15.3 & 65.4 & 30.2 & 39.8 & 22.8 & 19.2 & 12.2 & 17.1 & 36.9 & 1.7 & 21.4 \\
\bottomrule
\end{tabular}
}
\end{table*}

\paragraph{DataFlow shines in structured domains; direct LLM prompting is inconsistent.}
DataFlow-based methods tell a coherent story where data construction is more algorithmic. DataFlow-Skill takes the best Finance average on Qwen2.5-7B by a clear margin, and leads on Law as well, benefiting from a pipeline that processes every page of a textbook systematically. The limitation shows up in Science, where both DataFlow variants collapse on Llama-3.1-8B to well below the Dolly-only baseline. Expert-designed workflows are effective when the extraction task is regular; they struggle when the domain demands open-ended reasoning.

Direct LLM-based generation, by contrast, is the simplest baseline, and its performance reflects that simplicity. It can be competitive in some domains, but on Finance every direct generator falls below the Dolly-only baseline: when source books exceed the context window and there is no coverage tracking, a single-pass prompt is not enough.

\paragraph{Agents are the strongest overall family, especially in reasoning-heavy domains.}
In Math, Medical, and Law, agent-based generators claim most of the top slots. Multiple agent backbones cluster near the top of the Math leaderboard on Qwen2.5-7B, all ahead of every direct LLM and DataFlow competitor. In Medical, the best agent result beats all other families by a meaningful margin. The common thread is that agents can break a long document into chunks, decide what to extract from each, verify the output, and iterate, rather than processing the source in one shot. That flexibility matters most when the relevant knowledge is scattered across a lengthy text.

\paragraph{Skill-guided construction is the strongest method in knowledge-extraction-dense specialized domains.}
Data-Construction-Skill is designed for the setting in which the source is a dense domain book and the goal is to extract faithful, self-contained QA supervision. The Finance results on Llama-3.1-8B are the clearest validation: Skill improves the Dolly-only baseline by nearly 20 points absolute, a gain larger than any other method in that domain, and it holds its own in Medical as well. The skill layer (output schemas, filtering rules, coverage constraints) appears to be what makes the difference when faithfulness and completeness of extraction matter most.

\begin{table*}[!t]
\centering
\caption{Performance of Llama-3.1-8B on Science, Law, and Medical benchmarks after fine-tuning on datasets synthesized by different generators. All rows are fine-tuned on Dolly-15k (a general instruction-following corpus) jointly with the domain-specific dataset produced by the named generator; $^\dag$~marks the Dolly-15k only reference baseline, fine-tuned on Dolly-15k alone without any domain-specific synthetic data. Benchmark abbreviations: LB = LegalBench, LG = LexGLUE, MCR = MedCaseReasoning, MMCQA = MedMCQA, MRB = MedR-Bench, MSTEM = MMLU-STEM, MPRO = MMLU-Pro, SGPQA = SuperGPQA, CB = ChemBench, SB = SciBench.}
\label{tab:synthetic_data_slm_llama}
\resizebox{\textwidth}{!}{
\begin{tabular}{l|ccc|cccc|cccccccc}
\toprule
\multirow{2}{*}{\textbf{Training Data Generator}} 
& \multicolumn{3}{c|}{\textbf{Law}} 
& \multicolumn{4}{c|}{\textbf{Medical}} 
& \multicolumn{8}{c}{\textbf{Science}} \\
\cmidrule(lr){2-4}\cmidrule(lr){5-8}\cmidrule(lr){9-16}
& LB & LG & Avg 
& MCR & MMCQA & MRB & Avg 
& MSTEM & MPRO & GPQA & SGPQA & CB & PIQA & SB & Avg \\
\midrule
Dolly-15k only$^\dag$ & 87.9 & 51.1 & 69.5 & 9.9 & 17.1 & 34.3 & 20.4 & \textbf{25.3} & \textbf{14.7} & \textbf{14.1} & \textbf{8.2} & \textbf{14.7} & \underline{11.4} & \textbf{3.8} & \textbf{13.2} \\
\midrule
\rowcolor[rgb]{.867, .922, .969}
\multicolumn{16}{c}{\textit{\textbf{DataFlow-based Generators}}} \\
\midrule
DataFlow & 84.0 & 61.3 & 72.7 & 12.3 & 21.3 & 56.1 & 29.9 & 8.5 & 3.7 & 5.1 & 2.5 & 6.1 & 6.4 & 0.3 & 4.7 \\
DataFlow-Skill & 85.2 & 61.0 & 73.1 & 12.4 & \textbf{33.8} & 61.8 & \underline{36.0} & 5.6 & 1.7 & 3.3 & 1.7 & 4.6 & 4.8 & 0.0 & 3.1 \\
\midrule
\rowcolor[rgb]{.867, .922, .969}
\multicolumn{16}{c}{\textit{\textbf{LLM-based Generators}}} \\
\midrule
Claude Opus 4.6 & 85.9 & 56.1 & 71.0 & 16.3 & 27.2 & 64.8 & \textbf{36.1} & 17.1 & \underline{10.4} & 10.8 & \underline{6.1} & 9.2 & 9.6 & \underline{2.2} & \underline{9.3} \\
Gemini 3.0 Pro & 86.1 & 61.9 & 74.0 & 16.4 & 16.3 & 67.6 & 33.4 & 16.7 & 10.3 & 9.8 & \underline{6.1} & 10.2 & 9.2 & 1.6 & 9.1 \\
GPT-5.2 & \textbf{91.6} & 61.3 & \textbf{76.4} & 13.5 & 26.9 & 56.8 & 32.4 & 14.7 & 7.4 & 10.7 & 4.9 & 9.0 & \textbf{11.9} & 0.6 & 8.5 \\
\midrule
\rowcolor[rgb]{.867, .922, .969}
\multicolumn{16}{c}{\textit{\textbf{Agent-based Generators}}} \\
\midrule
Qwen3.5-Plus & 84.2 & \underline{62.3} & 73.2 & 16.7 & 13.7 & \underline{69.9} & 33.4 & \underline{17.3} & 9.5 & \underline{11.1} & 5.1 & \underline{12.1} & 8.9 & 0.9 & \underline{9.3} \\
GLM-4.7 & 83.8 & 60.7 & 72.2 & 15.6 & 20.0 & 67.1 & 34.2 & 16.9 & 9.9 & 9.1 & 5.4 & 12.0 & 8.4 & 0.6 & 8.9 \\
Claude Opus 4.6 & 87.4 & 52.1 & 69.8 & 14.8 & 18.4 & 64.8 & 32.7 & 17.0 & 9.4 & 10.7 & 5.7 & 10.9 & 7.8 & 1.9 & 9.1 \\
Gemini 3.0 Pro & 86.3 & 60.6 & 73.5 & 16.5 & 15.9 & \textbf{70.2} & 34.2 & 10.1 & 3.8 & 5.0 & 2.3 & 5.9 & 11.0 & 0.3 & 5.5 \\
GPT-5.2 & 84.5 & 61.3 & 72.9 & \underline{16.8} & 21.3 & 60.1 & 32.7 & 8.1 & 3.3 & 2.9 & 2.2 & 4.4 & 9.1 & 0.2 & 4.3 \\
GPT-5.3-codex & \underline{89.6} & \textbf{63.1} & \underline{76.3} & \textbf{18.3} & 23.7 & 65.0 & 35.7 & 16.4 & 9.0 & 10.4 & 5.3 & 8.5 & 7.7 & 0.6 & 8.3 \\
\midrule
\textbf{Skill~(Claude Opus 4.6)} & 86.6 & 53.7 & 70.2 & 13.7 & \underline{27.9} & 64.6 & 35.4 & 16.2 & 8.5 & 9.1 & 4.8 & 7.5 & 5.9 & 0.2 & 7.5 \\
\bottomrule
\end{tabular}
}
\end{table*}

\paragraph{Skill-guided construction has clear failure modes in science and open-ended reasoning domains.}
Skill does not win everywhere, and the pattern of its losses is informative. It falls behind the strongest agents on Medical and Law for Qwen2.5-7B, and on Science it trails the Dolly-only baseline on both base models. The losses cannot be blamed on a data-volume cap, since each method decides autonomously how much data to emit (Section~\ref{sec:data_construction_skill}). Two factors are likely at play: the Claude Opus 4.6 backbone is cautious on open-ended scientific reasoning, tending toward sparser outputs; and on Medical and Law, some agent backbones generate long case-style answers that happen to fit the benchmark format better than Skill's more structured extractions. In other words, Skill optimizes for faithfulness and atomic knowledge coverage, which helps on structured knowledge benchmarks but does not match benchmarks whose answer style rewards long, open-ended case narratives.

\subsection{Results of Data Quality Evaluation}
\label{sec:results_quality}

We report Pearson correlations between each metric's scores and empirical downstream performance, averaged across three base models (Qwen2.5-7B, Llama-3.1-8B, Mistral-7B-v0.3) per domain; higher positive correlation indicates stronger predictive alignment with downstream utility. We additionally track sign consistency, i.e., whether each metric's correlation has the theoretically expected sign, because a metric that produces the right ranking on average but flips sign in specific (domain, model) cells is not reliable in practice. Results are in Tables~\ref{tab:results-gms} and~\ref{tab:results-mfl}.

\paragraph{DAS is the strongest predictor in reasoning- and knowledge-intensive domains.}
DAS leads or ties for the lead in 4 of 6 domains, with particularly strong results in Math ($r > 0.93$ on two of the three base models, $p < 10^{-4}$) and Medical. Crucially, it is the only metric that exceeds 0.70 in Math, Science, and Medical simultaneously, the three domains where getting the distribution right seems to matter most. On Science, DAS weakens on one base model and falls below the significance threshold, which we attribute to the small candidate pool rather than a fundamental failure. The broader pattern is consistent with the theoretical motivation in Section~\ref{sec:das_theory}: when the downstream task demands specialized reasoning, the gap between a dataset's distribution and the target turns out to be a surprisingly direct predictor of what fine-tuning on it will do.

\begin{table*}[!t]
    \centering
    \small
    \renewcommand{\arraystretch}{1.2}
    \setlength{\tabcolsep}{3pt}
    \caption{
    Pearson correlation between data quality metric and downstream performance on the \textbf{General Text}, \textbf{Math}, and \textbf{Science} domains, along with computational overhead.
    \textbf{Time (s)} values represent the average processing duration per candidate dataset across all six domains of DataPrep-Bench (excluding model loading and target set embedding costs).
    Correlation values are reported with two decimal places.
    \colorbox{lightyellow}{Light yellow} and \colorbox{lightgreen}{light green} indicate positive and negative correlations, respectively. Underlined values in model columns indicate that the correlation is consistent with the theoretical expectation and reaches statistical significance ($p < 0.05$).
    In each domain, the strongest consistent average correlation in the Avg column is also bolded.
    }
    \resizebox{\textwidth}{!}{
    \begin{tabular}{l|c|cccc|cccc|cccc}
        \toprule
        \multirow{2}{*}{Algorithm} & \multirow{2}{*}{Time (s)} & \multicolumn{4}{c|}{General} & \multicolumn{4}{c|}{Math} & \multicolumn{4}{c}{Science} \\
        \cmidrule(lr){3-6} \cmidrule(lr){7-10} \cmidrule(lr){11-14}
        & & Qwen & Llama-8B & Mistral & Avg & Qwen & Llama-8B & Mistral & Avg & Qwen & Llama-8B & Mistral & Avg \\
        \midrule
        \rowcolor[rgb]{.867, .922, .969}
        \multicolumn{14}{c}{\textit{\textbf{Distribution Based}}} \\
        \midrule
        DAS & 306.02
        & \pnum{\underline{0.64}} & \pnum{\underline{0.65}} & \pnum{\underline{0.74}} & \pnum{\textbf{0.68}}
        & \pnum{\underline{0.72}} & \pnum{\underline{0.93}} & \pnum{\underline{0.94}} & \pnum{\textbf{0.86}}
        & \pnum{0.53} & \pnum{\underline{0.82}} & \pnum{\underline{0.80}} & \pnum{0.72} \\
        \midrule
        \rowcolor[rgb]{.867, .922, .969}
        \multicolumn{14}{c}{\textit{\textbf{Quality Based}}} \\
        \midrule
        Qurating-WritingStyle & \multirow{4}{*}{84.65}
        & \pnum{0.17} & \pnum{0.36} & \pnum{0.38} & \pnum{0.30}
        & \nnum{0.48} & \nnum{0.33} & \nnum{0.46} & \nnum{0.42}
        & \nnum{0.10} & \nnum{0.16} & \pnum{0.04} & \nnum{0.07} \\
        Qurating-Expertise &
        & \nnum{0.17} & \nnum{0.37} & \nnum{0.42} & \nnum{0.32}
        & \pnum{\underline{0.61}} & \pnum{\underline{0.76}} & \pnum{\underline{0.71}} & \pnum{0.69}
        & \pnum{0.56} & \pnum{\underline{0.76}} & \pnum{\underline{0.83}} & \pnum{\textbf{0.72}} \\
        Qurating-FactsTrivia &
        & \pnum{0.03} & \pnum{0.10} & \pnum{0.02} & \pnum{0.05}
        & \nnum{0.35} & \nnum{0.04} & \nnum{0.20} & \nnum{0.20}
        & \nnum{0.08} & \pnum{0.14} & \pnum{0.32} & \pnum{0.13} \\
        Qurating-Educational &
        & \nnum{0.03} & \pnum{0.17} & \pnum{0.13} & \pnum{0.09}
        & \nnum{0.23} & \pnum{0.04} & \nnum{0.10} & \nnum{0.10}
        & \pnum{0.01} & \pnum{0.17} & \pnum{0.34} & \pnum{0.18} \\
        \midrule
        BERTVendi & \multirow{2}{*}{459.37}
        & \pnum{0.38} & \pnum{0.47} & \pnum{\underline{0.58}} & \pnum{0.47}
        & \nnum{0.69} & \nnum{0.40} & \nnum{0.54} & \nnum{0.54}
        & \nnum{0.41} & \nnum{0.10} & \nnum{0.07} & \nnum{0.19} \\
        SimCSEVendi &
        & \pnum{0.30} & \pnum{0.46} & \pnum{\underline{0.55}} & \pnum{0.44}
        & \nnum{0.80} & \nnum{0.50} & \nnum{0.63} & \nnum{0.65}
        & \nnum{0.66} & \nnum{0.43} & \nnum{0.34} & \nnum{0.48} \\
        \midrule
        Deita-Quality & 434.99
        & \pnum{0.06} & \pnum{0.46} & \pnum{0.18} & \pnum{0.23}
        & \nnum{0.42} & \nnum{0.46} & \nnum{0.54} & \nnum{0.47}
        & \pnum{0.08} & \nnum{0.26} & \nnum{0.13} & \nnum{0.10} \\
        RewardModel & 109.61
        & \pnum{\underline{0.56}} & \pnum{\underline{0.67}} & \pnum{\underline{0.54}} & \pnum{0.59}
        & \nnum{0.71} & \nnum{0.81} & \nnum{0.85} & \nnum{0.79}
        & \nnum{0.39} & \nnum{0.54} & \nnum{0.32} & \nnum{0.42} \\
        Superfiltering & 40.05
        & \pnum{0.13} & \pnum{0.25} & \pnum{0.30} & \pnum{0.23}
        & \nnum{0.11} & \nnum{0.21} & \nnum{0.29} & \nnum{0.21}
        & \pnum{0.02} & \nnum{0.10} & \nnum{0.13} & \nnum{0.07} \\
        FineWebEdu & 11.67
        & \nnum{0.43} & \nnum{0.43} & \nnum{0.37} & \nnum{0.41}
        & \pnum{0.42} & \pnum{\underline{0.74}} & \pnum{\underline{0.65}} & \pnum{0.60}
        & \pnum{0.09} & \pnum{0.45} & \pnum{0.59} & \pnum{0.37} \\
        PairQual & 12.34
        & \nnum{0.03} & \nnum{0.03} & \nnum{0.15} & \nnum{0.07}
        & \nnum{0.16} & \pnum{0.06} & \nnum{0.06} & \nnum{0.05}
        & \pnum{0.19} & \pnum{0.23} & \pnum{0.40} & \pnum{0.27} \\
        Perplexity & 40.60
        & \pnum{0.47} & \pnum{0.43} & \pnum{0.48} & \pnum{0.46}
        & \nnum{0.71} & \nnum{0.56} & \nnum{0.62} & \nnum{0.63}
        & \nnum{0.59} & \nnum{0.32} & \nnum{0.44} & \nnum{0.45} \\
        \midrule
        \rowcolor[rgb]{.867, .922, .969}
        \multicolumn{14}{c}{\textit{\textbf{Diversity Based}}} \\
        \midrule
        MTLD & \multirow{2}{*}{2.59}
        & \pnum{0.48} & \pnum{0.38} & \pnum{0.32} & \pnum{0.39}
        & \nnum{0.58} & \nnum{0.59} & \nnum{0.67} & \nnum{0.62}
        & \nnum{0.48} & \nnum{0.67} & \nnum{0.49} & \nnum{0.55} \\
        HD-D &
        & \pnum{0.17} & \nnum{0.08} & \pnum{0.02} & \pnum{0.03}
        & \nnum{0.34} & \nnum{0.17} & \nnum{0.25} & \nnum{0.25}
        & \nnum{0.64} & \nnum{0.61} & \nnum{0.44} & \nnum{0.56} \\
        \midrule
        Task2Vec & 7.93
        & \nnum{0.04} & \pnum{0.16} & \pnum{0.17} & \pnum{0.10}
        & \nnum{0.48} & \nnum{0.09} & \nnum{0.25} & \nnum{0.27}
        & \nnum{0.65} & \nnum{0.36} & \nnum{0.21} & \nnum{0.41} \\
        Ngram & 1.06
        & \nnum{0.01} & \nnum{0.05} & \nnum{0.00} & \nnum{0.02}
        & \nnum{0.31} & \pnum{0.12} & \nnum{0.05} & \nnum{0.08}
        & \nnum{0.22} & \pnum{0.04} & \pnum{0.22} & \pnum{0.01} \\
        Deita-Complexity & 344.34
        & \pnum{\underline{0.52}} & \pnum{0.25} & \pnum{0.26} & \pnum{0.34}
        & \nnum{0.19} & \nnum{0.34} & \nnum{0.32} & \nnum{0.28}
        & \pnum{0.05} & \pnum{0.27} & \pnum{0.32} & \pnum{0.22} \\
        \bottomrule
    \end{tabular}
    \label{tab:results-gms}
    }
\end{table*}

\paragraph{Existing metrics are either narrow specialists or sign-inconsistent.}
No existing metric is reliable across domains. Each one has a domain or two where it looks reasonable and others where it falls apart, sometimes flipping sign entirely, meaning it would actively recommend worse data. QuRating-Expertise does well on Math and Science but collapses on General Text and Medical. Superfiltering leads on Finance but is nearly uninformative on Math and Science. BERTVendi is usable in General Text and little else. The sign-inconsistency is the more serious concern: a metric that ranks data correctly on average but flips in specific settings is actively misleading for practitioners who cannot know in advance which settings they are in. A single quality score cannot substitute for knowing where a dataset sits relative to the target domain.

\paragraph{Finance and Law are harder for all metrics, not just DAS.}
DAS is weaker in Finance and Law, which warrants explanation. Both candidate pools are structurally imbalanced: only one in-domain dataset sits alongside seven out-of-domain candidates, so most candidates are already distributionally far from the proxy and the metric has little variation to rank against. Beyond pool structure, Finance and Law benchmarks test highly specific professional conventions that generic text embeddings do not represent well. The key context is that \emph{no} metric in our evaluation clears a strong average correlation in these two domains; they stress all candidate-ranking metrics equally. We treat them as open challenges for the benchmark.

\begin{table*}[!t]
    \centering
    \small
    \renewcommand{\arraystretch}{1.2}
    \setlength{\tabcolsep}{3pt}
    \caption{
    Pearson correlation between data quality metric and downstream performance on the \textbf{Medical}, \textbf{Finance}, and \textbf{Law} domains, along with computational overhead.
    \textbf{Time (s)} values represent the average processing duration per candidate dataset across all six domains of DataPrep-Bench (excluding model loading and target set embedding costs).
    Correlation values are reported with two decimal places.
    \colorbox{lightyellow}{Light yellow} and \colorbox{lightgreen}{light green} indicate positive and negative correlations, respectively. Underlined values in model columns indicate that the correlation is consistent with the theoretical expectation and reaches statistical significance ($p < 0.05$).
    In each domain, the strongest consistent average correlation in the Avg column is also bolded.
    }
    \resizebox{\textwidth}{!}{
    \begin{tabular}{l|c|cccc|cccc|cccc}
        \toprule
        \multirow{2}{*}{Algorithm} & \multirow{2}{*}{Time (s)} & \multicolumn{4}{c}{Medical} & \multicolumn{4}{c}{Finance} & \multicolumn{4}{c}{Law} \\
        \cmidrule(lr){3-6} \cmidrule(lr){7-10} \cmidrule(lr){11-14}
        & & Qwen & Llama-8B & Mistral & Avg & Qwen & Llama-8B & Mistral & Avg & Qwen & Llama-8B & Mistral & Avg \\
        \midrule
        \rowcolor[rgb]{.867, .922, .969}
        \multicolumn{14}{c}{\textit{\textbf{Distribution Based}}} \\
        \midrule
        DAS & 306.02
        & \pnum{\underline{0.73}} & \pnum{\underline{0.87}} & \pnum{\underline{0.72}} & \pnum{\textbf{0.77}}
        & \nnum{0.18} & \pnum{0.57} & \pnum{0.14} & \pnum{0.18}
        & \nnum{0.50} & \nnum{0.17} & \nnum{0.40} & \nnum{0.36} \\
        \midrule
        \rowcolor[rgb]{.867, .922, .969}
        \multicolumn{14}{c}{\textit{\textbf{Quality Based}}} \\
        \midrule
        Qurating-WritingStyle & \multirow{4}{*}{84.65}
        & \nnum{0.10} & \pnum{0.20} & \nnum{0.06} & \pnum{0.01}
        & \pnum{0.03} & \pnum{0.13} & \pnum{0.33} & \pnum{0.16}
        & \pnum{\underline{0.63}} & \nnum{0.08} & \pnum{\underline{0.64}} & \pnum{0.40} \\
        Qurating-Expertise &
        & \pnum{0.56} & \pnum{\underline{0.73}} & \pnum{\underline{0.66}} & \pnum{0.65}
        & \nnum{0.04} & \pnum{0.13} & \pnum{0.60} & \pnum{0.23}
        & \pnum{\underline{0.72}} & \nnum{0.37} & \pnum{0.37} & \pnum{0.24} \\
        Qurating-FactsTrivia &
        & \pnum{0.31} & \pnum{\underline{0.66}} & \pnum{0.36} & \pnum{0.44}
        & \nnum{0.06} & \pnum{0.23} & \pnum{0.47} & \pnum{0.21}
        & \pnum{0.58} & \nnum{0.15} & \pnum{\underline{0.70}} & \pnum{0.37} \\
        Qurating-Educational &
        & \pnum{0.05} & \pnum{0.40} & \pnum{0.14} & \pnum{0.20}
        & \nnum{0.05} & \pnum{0.10} & \pnum{0.49} & \pnum{0.18}
        & \pnum{0.57} & \nnum{0.27} & \pnum{\underline{0.64}} & \pnum{0.31} \\
        Deita-Quality & 434.99
        & \pnum{0.57} & \pnum{\underline{0.63}} & \pnum{\underline{0.68}} & \pnum{0.63}
        & \nnum{0.18} & \nnum{0.45} & \pnum{0.15} & \nnum{0.16}
        & \pnum{0.46} & \nnum{0.59} & \pnum{0.24} & \pnum{0.04} \\
        RewardModel & 109.61
        & \pnum{0.26} & \pnum{0.32} & \pnum{0.23} & \pnum{0.27}
        & \nnum{0.07} & \nnum{0.30} & \nnum{0.06} & \nnum{0.14}
        & \pnum{0.43} & \nnum{0.08} & \pnum{0.51} & \pnum{0.28} \\
        Superfiltering & 40.05
        & \nnum{0.29} & \nnum{0.22} & \nnum{0.43} & \nnum{0.31}
        & \pnum{\underline{0.71}} & \pnum{0.60} & \pnum{\underline{0.69}} & \pnum{\textbf{0.67}}
        & \pnum{0.52} & \pnum{0.52} & \pnum{0.41} & \pnum{0.48} \\
        FineWebEdu & 11.67
        & \pnum{0.18} & \pnum{0.55} & \pnum{0.29} & \pnum{0.34}
        & \nnum{0.23} & \pnum{0.16} & \pnum{0.48} & \pnum{0.13}
        & \pnum{0.57} & \nnum{0.22} & \pnum{0.59} & \pnum{0.31} \\
        PairQual & 12.34
        & \pnum{0.44} & \pnum{\underline{0.71}} & \pnum{0.53} & \pnum{0.56}
        & \nnum{0.10} & \pnum{0.09} & \pnum{0.39} & \pnum{0.13}
        & \pnum{\underline{0.63}} & \nnum{0.28} & \pnum{0.61} & \pnum{0.32} \\
        Perplexity & 40.60
        & \nnum{0.53} & \nnum{0.30} & \nnum{0.64} & \nnum{0.49}
        & \pnum{0.56} & \pnum{0.59} & \pnum{0.35} & \pnum{0.50}
        & \pnum{0.06} & \pnum{0.33} & \pnum{0.53} & \pnum{0.31} \\
        BERTVendi & \multirow{2}{*}{459.37}
        & \nnum{0.64} & \nnum{0.26} & \nnum{0.56} & \nnum{0.49}
        & \pnum{0.33} & \pnum{\underline{0.79}} & \pnum{0.23} & \pnum{0.45}
        & \pnum{0.42} & \pnum{0.20} & \pnum{0.44} & \pnum{0.35} \\
        SimCSEVendi &
        & \nnum{0.47} & \nnum{0.09} & \nnum{0.50} & \nnum{0.35}
        & \pnum{0.13} & \pnum{0.54} & \pnum{0.16} & \pnum{0.28}
        & \pnum{0.28} & \pnum{0.15} & \pnum{\underline{0.74}} & \pnum{0.39} \\
        \midrule
        \rowcolor[rgb]{.867, .922, .969}
        \multicolumn{14}{c}{\textit{\textbf{Diversity Based}}} \\
        \midrule
        Task2Vec & 7.93
        & \pnum{0.06} & \pnum{0.39} & \nnum{0.03} & \pnum{0.14}
        & \nnum{0.26} & \pnum{0.20} & \pnum{0.29} & \pnum{0.08}
        & \pnum{0.12} & \nnum{0.12} & \pnum{\underline{0.91}} & \pnum{0.30} \\
        MTLD & \multirow{2}{*}{2.59}
        & \pnum{0.27} & \pnum{0.48} & \pnum{0.23} & \pnum{0.33}
        & \pnum{0.17} & \pnum{0.47} & \pnum{0.37} & \pnum{0.34}
        & \pnum{\underline{0.71}} & \pnum{0.26} & \pnum{0.59} & \pnum{\textbf{0.52}} \\
        HD-D &
        & \pnum{0.20} & \pnum{0.55} & \pnum{0.16} & \pnum{0.30}
        & \pnum{0.15} & \pnum{0.48} & \pnum{0.40} & \pnum{0.34}
        & \pnum{0.45} & \pnum{0.03} & \pnum{\underline{0.84}} & \pnum{0.44} \\
        Ngram & 1.06
        & \pnum{0.12} & \pnum{0.54} & \pnum{0.18} & \pnum{0.28}
        & \nnum{0.02} & \pnum{0.50} & \pnum{0.42} & \pnum{0.30}
        & \pnum{0.54} & \nnum{0.01} & \pnum{\underline{0.66}} & \pnum{0.40} \\
        Deita-Complexity & 344.34
        & \nnum{0.10} & \nnum{0.02} & \nnum{0.09} & \nnum{0.07}
        & \pnum{0.57} & \pnum{0.06} & \pnum{0.44} & \pnum{0.36}
        & \pnum{0.55} & \pnum{0.01} & \pnum{0.25} & \pnum{0.27} \\
        \bottomrule
    \end{tabular}
    \label{tab:results-mfl}
    }
\end{table*}

\paragraph{DAS offers the best trade-off between accuracy and compute among the stronger metrics.}
We include wall-clock processing time in Table~\ref{tab:results-gms}. DAS is roughly $1.4$--$1.5\times$ faster than the two comparable high-accuracy baselines (BERTVendi and Deita-Quality), which are both slower and less accurate. Cheap heuristics are much faster but their correlations on reasoning-heavy domains lag substantially. The practical summary: DAS is not the fastest option, but it is the fastest option that actually works.

\section{Conclusion}

In this work, we introduced \textbf{DataPrep-Bench}, a unified benchmark for LLM-driven training data preparation. DataPrep-Bench covers two complementary tracks: \textbf{Data Construction}, which evaluates whether LLMs, agents, and workflows can transform raw domain sources into useful supervised training data, and \textbf{Data Quality Evaluation}, which evaluates whether a scoring function can predict the downstream training value of candidate datasets. By grounding both tracks in end-to-end downstream performance, DataPrep-Bench moves beyond surface-level quality proxies and provides a faithful framework for studying data preparation end-to-end rather than through isolated case studies.

Our experiments reveal two complementary insights. First, adding synthesized domain data on top of the Dolly-15k instruction-following corpus frequently degrades the Dolly-only baseline across construction methods and model architectures, a finding that surface-level quality metrics would not have anticipated, underscoring the necessity of end-to-end, downstream-grounded evaluation. Within this landscape, \textbf{Data-Construction-Skill} provides a strong specialized baseline for knowledge-extraction-dense domains (Finance, Medical), where it lifts the Dolly-only baseline substantially and is competitive with the strongest agent- and DataFlow-based methods, suggesting that reusable skill-level specifications are a promising direction for corpus-level data preparation, though domain-specific design remains an open challenge. Second, our MMD-based \textbf{Distributional Alignment Score (DAS)} achieves the strongest cross-model correlation with downstream performance in four of six domains and is the only metric that clears $r>0.70$ simultaneously in Math, Science, and Medical. This confirms that distributional proximity to the target domain is a principled predictor of dataset utility, while also revealing that no single metric is uniformly reliable across all domains, with Finance and Law standing out as open challenges for every evaluated metric.

DataPrep-Bench provides a common testbed for future research on agentic data construction, automated data selection, and downstream-grounded evaluation of training corpora. In future work, we plan to (i)~expand the candidate pools for Finance, Law, and Medical as more high-quality in-domain SFT corpora become public; (ii)~investigate skill-guided designs tailored to open-ended scientific reasoning, where Skill currently underperforms; and (iii)~pair DAS with complementary evaluation signals (e.g., reference-free quality scoring or domain-aware embedding models) to improve predictive reliability in Finance and Law.

\clearpage

\bibliographystyle{plainnat}
\bibliography{main}

\clearpage

\beginappendix

\section{Detailed Introduction to the Benchmarks}
\label{app:benchmarks}

The following provides a detailed description of the benchmarks used in DataPrep-Bench.

\paragraph{General Text} We evaluated models on the following benchmark:
\begin{itemize}
    \item \textbf{MMLU-Redux}: A refined benchmark comprising 5,700 manually re-annotated questions across all 57 MMLU subjects, designed to correct ground truth errors and ambiguities in the original benchmark for more reliable evaluation of LLM multitask understanding.
\end{itemize}

\paragraph{Mathematical Reasoning} We evaluated models on several representative benchmarks using the Qwen2.5-Math evaluation framework~\cite{yang2024qwen2}:
\begin{itemize}
    \item \textbf{AIME 2024}: A challenging benchmark of competition-level mathematics problems from the 2024 American Invitational Mathematics Examination, designed to test advanced mathematical reasoning and problem-solving abilities.
    \item \textbf{AMC 2023}: A benchmark derived from the 2023 American Mathematics Competitions, consisting of multiple-choice problems that evaluate mathematical knowledge and reasoning across various topics.
    \item \textbf{Gaokao 2024}: A benchmark based on mathematics questions from China's 2024 National College Entrance Examination, used to assess LLM performance on advanced high-school-level mathematical problems.
    \item \textbf{GSM8K}: A dataset of grade school math word problems requiring multi-step reasoning, widely used to evaluate foundational mathematical problem-solving capabilities of language models.
    \item \textbf{MATH}: A benchmark of challenging competition mathematics problems spanning multiple domains, designed to measure advanced mathematical reasoning and problem-solving abilities.
    \item \textbf{MinervaMath}: A collection of quantitative reasoning problems drawn from various STEM sources, evaluating the ability of LLMs to solve complex technical and scientific mathematical problems.
    \item \textbf{OlympiadBench}: A bilingual multimodal benchmark comprising olympiad-level mathematics and science problems, designed to evaluate AGI-level mathematical and scientific reasoning capabilities.
\end{itemize}

\paragraph{Science Knowledge and Reasoning} We evaluated models on several benchmarks across different subdomains of science using the evaluation protocol of MegaScience~\cite{fan2025megascience}:
\begin{itemize}
    \item \textbf{MMLU-STEM}: The STEM-related subsets of MMLU, a massive multitask benchmark designed to evaluate broad scientific knowledge and reasoning across diverse technical domains.
    \item \textbf{MMLU-Pro}: A more robust and challenging extension of MMLU featuring an increased number of answer choices per question, designed to reduce guessing and better assess deep knowledge understanding.
    \item \textbf{GPQA}: A graduate-level Q\&A benchmark comprising difficult science questions written by domain experts, designed to be ``Google-proof'' and evaluate deep expert reasoning in scientific disciplines.
    \item \textbf{SuperGPQA}: A comprehensive benchmark scaling graduate-level evaluation across 285 disciplines, employing human-LLM collaborative filtering to assess expert-level scientific knowledge and reasoning capabilities.
    \item \textbf{ChemBench}: A chemistry-focused benchmark designed to evaluate whether LLMs possess expert-level chemical knowledge and reasoning abilities across diverse problem types.
    \item \textbf{PIQA}: A benchmark for reasoning about physical commonsense in natural language, testing the ability of models to understand and apply intuitive physics principles.
    \item \textbf{SciBench}: A collection of college-level scientific problem-solving tasks spanning multiple disciplines, designed to evaluate advanced reasoning and calculation abilities in scientific contexts.
\end{itemize}

\paragraph{Specialized Medical Expertise} We evaluated models on three widely used medical benchmarks:
\begin{itemize}
    \item \textbf{MedR-Bench}: A reasoning-focused benchmark of 1,453 structured real-world clinical cases derived from PubMed Central case reports, covering 13 body systems and 10 medical specialties. It is designed to evaluate the clinical reasoning performance of LLMs.
    \item \textbf{MedMCQA}: A large-scale multiple-choice question-answering dataset with over 194,000 medical entrance exam questions across 21 subjects and ~2,400 healthcare topics, used to assess medical knowledge and reasoning.
    \item \textbf{MedCaseReasoning}: A clinical diagnostic QA benchmark composed of thousands of diagnostic pairs extracted from clinical case reports, evaluating the ability of models to produce correct diagnoses based on detailed case information.
\end{itemize}

\paragraph{Specialized Financial Expertise} We selected three benchmarks representing different aspects of financial reasoning and knowledge:
\begin{itemize}
    \item \textbf{XFinBench}: A benchmark designed to evaluate LLMs' ability to solve complex, knowledge-intensive financial problems across graduate-level finance topics, including statement judgment, multiple-choice questions, and financial calculation tasks.
    \item \textbf{FinEval-KR}: A framework and dataset for assessing LLMs' financial knowledge and reasoning, providing decoupled metrics to analyze performance across 22 financial subfields.
    \item \textbf{CPA-KQA}: A dataset derived from CPA examination materials, containing multiple-choice questions that measure detailed financial and accounting concept understanding and reasoning skills.
\end{itemize}

\paragraph{Specialized Legal Expertise} We evaluated models on two legal domain benchmarks:
\begin{itemize}
    \item \textbf{LegalBench}: A collaboratively constructed benchmark for measuring legal reasoning capabilities of LLMs, consisting of tasks designed by legal experts to reflect realistic legal reasoning challenges.
    \item \textbf{LexGLUE}: A standard legal benchmark comprising multiple legal natural language understanding tasks, including judgment prediction and contract analysis, to evaluate broad linguistic and reasoning capabilities in legal contexts.
\end{itemize}

\begin{table*}[t]
\centering
\caption{Number of training samples synthesized by each data construction method across six domains, all operating on the \emph{same} raw source corpus. Because the input material is held constant, data yield serves as a proxy for each method's \emph{extraction efficiency}---its capacity to convert raw domain content into instructional QA samples. DataFlow-based pipelines achieve the highest overall yield, while direct LLM-based generators produce substantially fewer samples due to context-window constraints. Agent-based methods exhibit pronounced domain-level variation, suggesting selective extraction behavior shaped by the underlying model. The Total column aggregates yield across all six domains.}
\label{tab:dataset-sizes}
\resizebox{0.8\textwidth}{!}{%
\begin{tabular}{l|rrrrrr|r}
\toprule
\textbf{Generator} & \textbf{Math} & \textbf{General} & \textbf{Finance} & \textbf{Law} & \textbf{Medicine} & \textbf{Science} & \textbf{Total} \\
\midrule
\rowcolor[rgb]{.867,.922,.969}
\multicolumn{8}{c}{\textit{\textbf{DataFlow-based Generators}}} \\
\midrule
DataFlow        & 90{,}222  & 102{,}975 & 111{,}774 &  53{,}891 & 253{,}355 & 133{,}190 & 745{,}407 \\
DataFlow-Skill  & 165{,}345 & 167{,}777 &  49{,}841 &  21{,}659 & 113{,}305 & 331{,}366 & 849{,}293 \\
\midrule
\rowcolor[rgb]{.867,.922,.969}
\multicolumn{8}{c}{\textit{\textbf{LLM-based Generators}}} \\
\midrule
Claude Opus 4.6 &  2{,}563  &   1{,}550 &   3{,}750 &   2{,}184 &   7{,}677 &   3{,}269 &  20{,}993 \\
Gemini 3.0 Pro  &  1{,}619  &   1{,}493 &   1{,}584 &     893   &   2{,}992 &   3{,}441 &  12{,}022 \\
GPT-5.2         &  9{,}179  &   3{,}473 &   5{,}832 &   3{,}949 &  14{,}945 &  18{,}238 &  55{,}616 \\
\midrule
\rowcolor[rgb]{.867,.922,.969}
\multicolumn{8}{c}{\textit{\textbf{Agent-based Generators}}} \\
\midrule
Qwen3.5-Plus    &     684   &   2{,}668 &  21{,}693 &  13{,}712 &  25{,}965 &  80{,}610 & 145{,}332 \\
GLM-4.7         &     684   &   2{,}668 &  23{,}246 &  13{,}326 &  24{,}734 &  50{,}552 & 115{,}210 \\
Claude Opus 4.6 & 14{,}892  & 311{,}950 &  17{,}059 &  19{,}505 &  44{,}284 &   2{,}480 & 410{,}170 \\
Gemini 3.0 Pro  &  5{,}340  &   1{,}188 &     249   &     389   &     465   &  56{,}987 &  64{,}618 \\
GPT-5.2         & 14{,}337  &   2{,}357 &  11{,}999 &   4{,}967 &  30{,}765 & 103{,}096 & 167{,}521 \\
GPT-5.3-codex   &  2{,}500  &  28{,}356 &     249   &   1{,}584 &     463   &   2{,}547 &  35{,}699 \\
\midrule
\textbf{Skill~(Claude Opus 4.6)} &  1{,}439  &   4{,}336 & 106{,}401 &  24{,}255 &  38{,}351 & 352{,}663 & 527{,}445 \\
\bottomrule
\end{tabular}%
}
\end{table*}

\section{Implementation Details}
\label{app:impl}

\subsection{Data Construction: Training Hyperparameters}

We fine-tune each base model on each constructed dataset with LlamaFactory~\cite{zheng2024llamafactory}, using a consistent set of hyperparameters across all data construction experiments to ensure comparability. Specifically, all models are fine-tuned for 3 epochs using a cosine learning rate scheduler with an initial learning rate of $5.0 \times 10^{-6}$ and a warmup ratio of 0.1. We employ a \texttt{per\_device\_train\_batch\_size} of 1 and \texttt{gradient\_accumulation\_steps} of 4, resulting in a total global batch size of 32 across the 8$\times$H20 GPU cluster.

We follow standard evaluation protocols for each benchmark. For Math, we use a decoding temperature of $0.6$ and a $16{,}384$ token context window. For General Text (MMLU-Redux), we employ a 5-shot prompting strategy. Domain-level performance is aggregated by averaging the scores across all benchmarks within each domain.

We use Qwen2.5-7B~\cite{qwen2.5} and Llama-3.1-8B~\cite{grattafiori2024llama} as base models for data construction experiments. Since these models have not undergone instruction-following training, Dolly-15k\footnote{\url{https://huggingface.co/datasets/databricks/databricks-dolly-15k}} is used as a general instruction-following corpus in all fine-tuning runs. Concretely, every row labelled with a specific construction method is fine-tuned on Dolly-15k jointly with the domain-specific constructed data, while the Dolly-15k only row (marked $^\dag$ in each table) is fine-tuned on Dolly-15k exclusively, with no domain-specific synthetic data. This design isolates the downstream effect of domain data construction from the confound of basic instruction-following ability.

\subsection{Dataset Size Design}

We intentionally do not impose a uniform cap on the number of synthesized samples across construction methods. Instead, each method autonomously decides how much data to emit from the same raw source corpus: DataFlow-based methods emit whatever their pipeline produces, direct LLM-based generators emit whatever fits in their context windows, and agent-based (including Skill) generators continue until the agent itself judges the construction to be complete. This reflects how such methods are deployed in practice (practitioners typically run a method to completion rather than truncating its output to a fixed budget) and it treats data volume efficiency as an intrinsic property of a construction method rather than a nuisance variable to be controlled away. One consequence is that higher-yield agents receive more fine-tuning signal than lower-yield DataFlow pipelines, so the downstream scores in Tables~\ref{tab:synthetic_data_mgf_qwen}--\ref{tab:synthetic_data_slm_llama} reflect the combined effect of construction quality and construction yield.

The resulting per-method, per-domain sample counts are reported in Table~\ref{tab:dataset-sizes}. Several patterns are worth noting. First, DataFlow-based pipelines achieve the highest overall yield (745K--849K samples), owing to their multi-step extraction design that exhaustively processes every page of input material. Second, direct LLM-based generators produce an order of magnitude fewer samples (12K--56K total), as each book is processed in a single pass bounded by the model's context window. Third, agent-based generators show pronounced domain-level variation: for instance, Claude Opus 4.6 extracts the majority of its samples from General Text (311K out of 410K total), while Skill concentrates heavily on Science and Finance, suggesting that the underlying model's generation tendencies, rather than the raw source size alone, shape the final distribution. These observations highlight that data construction method choice involves an implicit trade-off between extraction breadth and domain focus.

\subsection{Data Quality Evaluation: Implementation Details}

To evaluate whether data quality metrics generalize across model architectures, we conduct experiments with three representative base models: Qwen2.5-7B~\cite{qwen2.5}, Llama-3.1-8B~\cite{grattafiori2024llama}, and Mistral-7B-v0.3~\cite{jiang2023mistral}. These models span distinct architectural families, allowing us to test whether the observed correlation between distributional alignment and downstream performance is model-dependent or more broadly applicable. We use the same fine-tuning and benchmark evaluation protocol as in the Data Construction experiments unless otherwise specified.

For DAS computation, we encode dataset samples into 4096-dimensional embeddings using Qwen3-Embedding-8B. For each dataset, we randomly sample 5{,}000 instances to balance statistical stability and computational efficiency. We compute MMD using a Gaussian RBF kernel with fixed bandwidth $\sigma = 1.0$. The same sampling size, embedding model, and kernel configuration are used across all domains and candidate datasets.

\section{Prompts}

\subsection{Prompt of LLM-based Generators}

\begin{figure}[t]
    \centering
    \includegraphics[width=0.6\textwidth]{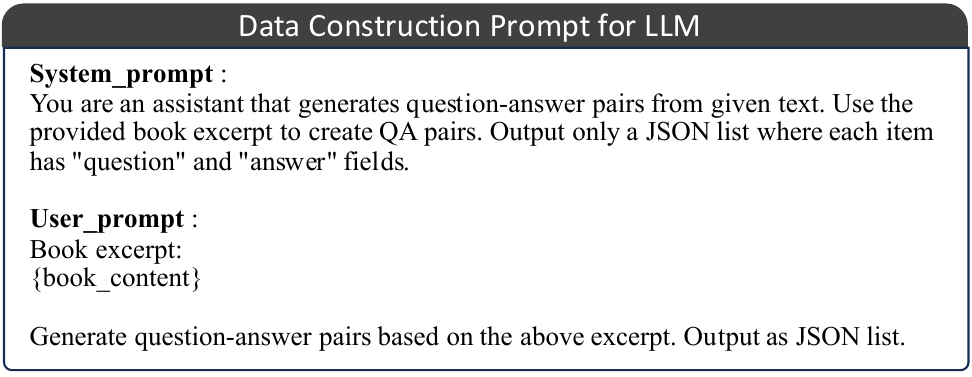}
    \caption{Data construction prompt used to guide the LLM.}
    \label{fig:prompt4llm}
\end{figure}

Figure~\ref{fig:prompt4llm} shows the prompt used to guide the LLM during the data construction process.

\subsection{Prompt of Agent-based Generators}

\begin{figure}[t]
    \centering
    \includegraphics[width=0.6\textwidth]{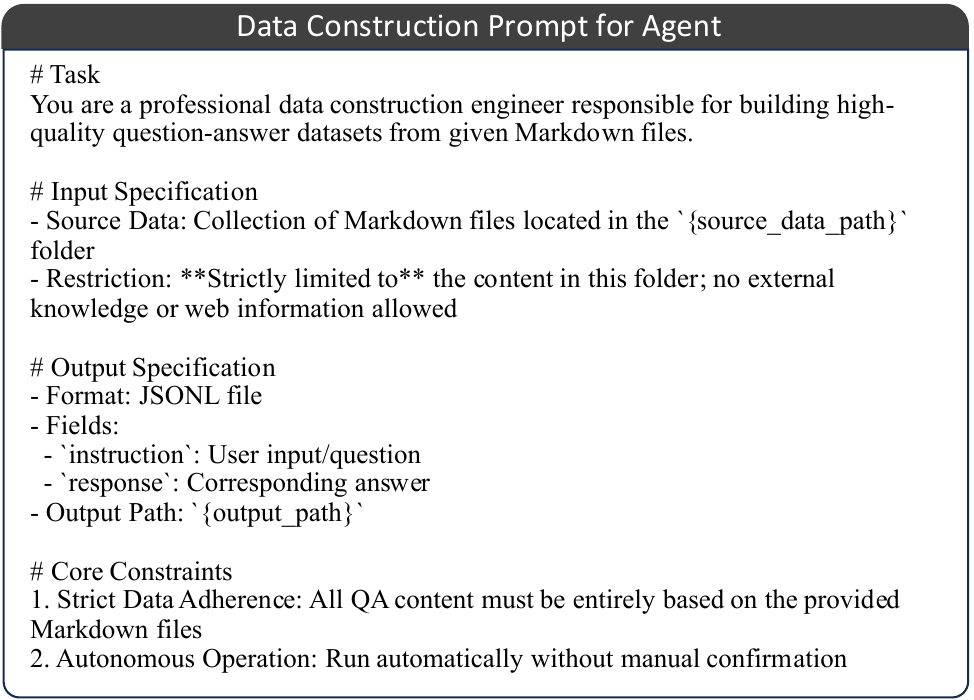}
    \caption{Data construction prompt used to guide the agent.}
    \label{fig:prompt4agent}
\end{figure}

Figure~\ref{fig:prompt4agent} shows the prompt used to guide the agent during the data construction process.
The prompt specifies the overall task objective, the expected input format, and the required output format.
It instructs the agent to extract reusable knowledge from source documents and transform it into structured supervision data.

\subsection{Prompt of Agent-based Generators with a Skill}

For the data construction agents that employ a skill (see Section~\ref{sec:data_construction_skill}), we adopt the same prompt shown in Figure~\ref{fig:prompt4agent}. Before the agent starts execution, the skill is automatically loaded and provides detailed procedural instructions and quality constraints.
This skill defines how the agent should identify knowledge-bearing chunks, generate different types of supervision samples, and validate the generated data.
Specifically, it supports three sample types: \texttt{concept\_qa}, \texttt{process\_qa}, and \texttt{case\_application}.
It also specifies the JSON schema, chunk-level status tracking, reasoning pattern constraints, and the quality rubric used during data construction.

Due to space limitations, the full content of the data construction skill is omitted from the main text and is available in the Clawhub project repository\footnote{\url{https://clawhub.ai/technomad-ds/data-construction-skill}}. To use the skill, one can download it via the repository's CLI command and place it in the directory expected by the target code agent (e.g., Claude Code). After providing the path to the raw document, the code agent automatically invokes the skill and initiates data construction.

\end{document}